\documentclass{article}

\usepackage[utf8]{inputenc}
\usepackage[T1]{fontenc}

\usepackage[preprint]{icml2026}
\usepackage{hyperref}
\hypersetup{hidelinks}

\usepackage{booktabs}
\usepackage{amsfonts}
\usepackage{amsmath}
\usepackage{amssymb}
\usepackage{nicefrac}
\usepackage{microtype}
\usepackage{tikz}
\usepackage{pgfplots}
\pgfplotsset{compat=1.18}
\usepgfplotslibrary{groupplots}
\usepackage{enumitem}
\usepackage{array}
\usetikzlibrary{positioning,arrows.meta,fit,calc,backgrounds}

\icmltitlerunning{Doc-to-LoRA Analysis: Query Conditioning and Interference}

\begin{document}

\twocolumn[
\icmltitle{Doc-to-LoRA Analysis: Query Conditioning, Wrong Facts, and Distractor Interference}

\icmlsetsymbol{equal}{*}

\begin{icmlauthorlist}
\icmlauthor{Jesper Derehag}{midbrain}
\end{icmlauthorlist}

\icmlaffiliation{midbrain}{Midbrain}

\icmlcorrespondingauthor{Jesper Derehag}{jesper@midbrain.ai}

\printAffiliationsAndNotice{}
]

{\let\thefootnote\relax\footnotetext{\hspace*{-\footnotesep}%
\textsuperscript{1}\,Midbrain. Correspondence to: Jesper Derehag
\textless{}jesper@midbrain.ai\textgreater{}.}}

\begin{abstract}
Doc-to-LoRA (D2L) compiles documents into low-rank adapters, allowing language
models to use their contents without repeatedly processing the original text.
We investigate its suitability for internalizing user memories through factual
question answering, controlled interventions, and a production MidBrain memory
pipeline. Across Gemma-2-2B and Qwen3-4B, D2L loses accuracy relative to
in-context learning, with additional documents introducing substantial
interference. Single-document errors frequently match answers to other questions
about the same passage, suggesting unreliable question--answer attribution.
Layer-wise interventions show that interference accumulates across depth,
with more concentrated early damage at larger bank sizes on Gemma. With eight
documents, applying the full bank in the initial layers gives similar accuracy
to withholding all document contributions there. Source membership is
detectable in intermediate states, but trained readers do not reliably
resolve the interference.
On PersonaMem-v2, joint compilation and normalized adapter composition achieve
accuracy close to consolidated text memory when retrieved episodes remain text,
but the resulting token savings are modest. Measured throughput gains also
depend on model and batch size despite substantial reductions in analytical
compute. Together, these findings limit the practical case for D2L as a
replacement for text-based memory and motivate investigating online-learning
methods that acquire personalized behavior from interactions in which memory
is used.
\end{abstract}

\section{Introduction}
\label{sec:intro}

Our memory system, MidBrain\footnote{\url{https://midbrain.ai}}, gives each
agent--user pair a consolidated semantic and procedural memory, derived from episodic
memory that is itself retrieved per query~\citep{tulving1972episodic,derehag2026smartsearch}.
In MidBrain's production system, the consolidated memory is injected as part of the
context. The same consolidated memory is added to the majority of requests, and the
model repeatedly processes that text during inference. We therefore
investigate whether this memory can be internalized, that is, made part of the model
rather than part of the context.

Doc-to-LoRA (D2L)~\citep{charakorn2026doctolora} is a candidate for this
internalization. A hypernetwork, meta-trained to approximate context
distillation~\citep{snell2022learning}, maps a document to a rank-8
adapter~\citep{hu2022lora} in one forward pass, and the adapter is installed in the
model in place of the document. When several documents are supplied, D2L adds their
adapters before the model processes the request. This composition rule is useful when
chunks of one document should contribute together, yet it also means that an adapter
can affect a request even when its document is irrelevant. The released compilers
recover much of the in-context accuracy on single factual
documents, while accuracy collapses when several
document adapters are summed (Section~\ref{sec:res-gap}).

We evaluate D2L as the internalization mechanism for MidBrain. The analysis
asks where document-conditioned computation is lost, how composition introduces
interference, and what these failures imply for using compiled user memory.
It addresses three questions.
\begin{enumerate}[label=\textbf{RQ\arabic*},leftmargin=*,nosep]
\item Can D2L adapters reach in-context
learning (ICL) parity when the memory is a single document?
\begin{enumerate}[label=\textbf{RQ1.\arabic*},leftmargin=*,nosep]
\item Is document content detectable in the compiler latent, and what passage
content appears in incorrect answers?
\item Can a different readout, compiler fine-tuning, or a write fitted to one passage
close the gap?
\item Which part of the in-context computation does the adapter fail to supply?
\end{enumerate}
\item Can D2L selectively ignore irrelevant documents in the way attention does
when the memory consists of several documents?
\begin{enumerate}[label=\textbf{RQ2.\arabic*},leftmargin=*,nosep]
\item How much of the composition loss can uniform update normalization recover?
\item In which layers do the distractors do their damage, and in which does the source
do its useful work?
\item Can the model's state in those layers tell the source from the distractors?
\end{enumerate}
\item Can compiled memory provide composable user memory with a practical
serving-cost advantage over text?
\end{enumerate}

\textbf{This paper makes the following contributions:}

\begin{enumerate}[leftmargin=*,nosep]
\item We characterize single-document errors: 46\% of incorrect D2L responses
on teacher-correct questions match another answer from the same passage.
Readout and fitting controls leave a gap to ICL at the tested budgets
(Section~\ref{sec:res-single}).
\item We localize composition damage across model depth. On Gemma at $K=8$, applying
the full bank below layer 8 gives .611 F1, close to .633 when those layers
receive no document contributions; at $K=16$, the former falls to .111.
The interventions distinguish accumulated interference from the cost of
withholding source information (Figure~\ref{fig:layers}).
\item We assess compiled production memory and serving cost. Joint compilation
gives .391 accuracy against .390 for text on PersonaMem-v2, but throughput
advantages depend on model and batch size. These measurements constrain the
practical case for replacing text memory (Table~\ref{tab:user-memory-summary}
and Figure~\ref{fig:serving-tradeoff}).
\end{enumerate}

\section{Related Work}
\label{sec:related}

\textbf{Context distillation and generated adapters.} Context distillation trains a
model without a context to match the predictions it makes with that context
available~\citep{askell2021general,snell2022learning}. Prior work applies this idea
to fixed inputs, entity updates, and prompts~\citep{choi2023fixed,padmanabhan2023propagating,bhargava2024prompt}.
For knowledge injection, SELF-PARAM, Knowledge Modules, and prompt distillation train
a student for each context using generated question--answer pairs, a
document-conditioned teacher's hidden states, or the prompted model
itself~\citep{wang2025selfupdatable,caccia2025training,kujanpaa2025efficient}.
Hypernetworks remove the per-context optimization: HyperTuning and HINT generate
modules from examples or instructions~\citep{phang2023hypertuning,ivison2023hint},
while Generative Adapter maps a context to low-rank adapters
\citep{chen2025generative}. Text-to-LoRA maps task descriptions to adapters
\citep{charakorn2025texttolora}; D2L maps documents by meta-learning context
distillation~\citep{charakorn2026doctolora}; and SHINE reuses the frozen model's
layers as the hypernetwork~\citep{liu2026shine}. We use the released D2L compilers
and leave the hypernetwork frozen except in the compiler-tuning controls.

\textbf{Analyses of adapters as memory.} LoRA capacity has been characterized as a
function of rank and sequence length~\citep{xu2026lora}, while another study finds
that adapters restricted to the first half of the layers and to feed-forward weights
store facts better than late-layer or attention-only adapters on Llama-3.1-8B
\citep{back2026understanding}. Mostly Reading~\citep{pulici2026mostlyreading}
argues that open-book scores mostly measure a reading behavior shared by all
adapters, and uses cloze (fill-in-the-blank) probes~\citep{taylor1953cloze} to
isolate stored content; merges of sixteen or more adapters keep 87--89\% of the
open-book advantage and 12--16\% of the cloze advantage. COIN shows that knowledge
learned by next-token prediction becomes dependent on its preceding context
\citep{zhou2026coin}. Those studies analyze trained adapters. We analyze generated
adapters with a fixed compiler and use exact interventions on the composed state to
localize failure.

\textbf{Work building on D2L.} The Override Gap~\citep{cheng2026override} attributes
D2L's failure under knowledge conflict to an adapter margin that is roughly constant
across documents and improves accuracy by scaling the adapter in selected layers.
Our normalization control measures how much composition loss can be recovered
by uniformly scaling document updates. Doc-to-Atom compiles typed
atoms into micro-LoRAs with retrieval keys and assembles a query-specific adapter
from the top-ranked atoms, attributing D2L's errors to interference from irrelevant
content in one monolithic adapter~\citep{diao2026doctoatom}. Its composition
retrieves over a key bank that grows with memory and applies the selection before the
prompt is processed. The oracles in Section~\ref{sec:res-layers} measure the ceiling
of source selection from layer zero, while the probes measure source
membership information in hidden states. Zuo et al.\ compare document adapters
with compressed KV caches~\citep{zuo2026rethinking}; TMEM updates LoRA weights online
within an episode~\citep{ren2026scaling}, and RPMem and DPPM build user memory from
D2L-style compilers trained on each benchmark's own
questions~\citep{zhao2026rpmem,chen2026dppm}. These works leave the location of
composition failure within the network unresolved.

\textbf{Knowledge editing, adapter composition, context compression.} Feed-forward
layers act as key--value memories~\citep{geva2021transformer}, and ROME, MEMIT,
MEND, and WISE write or route facts into them
\citep{meng2022locating,meng2023massediting,mitchell2022fast,wang2024wise}; D2L
adapts the same MLP down-projections while taking free text as input. Task arithmetic,
TIES-Merging, and LoraHub combine trained adapters by addition, sign resolution, or
searched weights~\citep{ilharco2023editing,yadav2023tiesmerging,huang2024lorahub}.
D2L sums independently generated adapters without weights, and the readers we
analyze reweight that sum per rank. Gisting, MEND, and Cartridges compress a context
into tokens or a trained KV prefix~\citep{mu2023learning,li2024mend,eyuboglu2026cartridges};
prompt compression and KV eviction reduce the positions attended
\citep{jiang2023llmlingua,pan2024llmlingua2,zhang2023h2o,li2024snapkv}.
Our serving comparison uses full-prefill ICL.

\textbf{Agent memory and user-memory benchmarks.} Agent memory systems manage
different context tiers and consolidate facts over long histories
\citep{packer2023memgpt,chhikara2025mem0,xu2025amem,maharana2024evaluating,wu2025longmemeval}.
PersonaMem-v2 evaluates implicit preference inference over 32K-token histories
\citep{jiang2025personamemv2}. We use it to measure the utility of compiled state,
with MidBrain serving as a fixed memory provider.

\section{Methods}
\label{sec:methods}

\subsection{Compiled representation and composition}
\label{sec:d2l}

\begin{figure*}[t]
\centering
\resizebox{0.92\textwidth}{!}{%
\begin{tikzpicture}[font=\footnotesize,>={Stealth[length=4pt]},
 box/.style={draw,rounded corners=2pt,align=center,inner sep=3pt,line width=0.5pt},
 input/.style={box,fill=orange!12,draw=orange!60!black},
 frozen/.style={box,fill=gray!10,draw=gray!70},
 trained/.style={box,fill=violet!10,draw=violet!60},
 gen/.style={box,fill=teal!12,draw=teal!70!black},
 flow/.style={->,draw=gray!70!black,line width=0.6pt},
 tap/.style={->,draw=teal!60!black,line width=0.6pt},
 rowlabel/.style={anchor=west,font=\footnotesize\bfseries,text=gray!70!black}]
\def\xa{3.6}\def\xb{7.3}\def\xc{11.9}
\node[rowlabel] at (-1.2,1.05) {Write time (once per document)};
\node[input,minimum width=2.2cm,minimum height=1.0cm] (d3) at (0.24,-0.16) {};
\node[input,minimum width=2.2cm,minimum height=1.0cm] (d2) at (0.12,-0.08) {};
\node[input,minimum width=2.2cm,minimum height=1.0cm] (doc) at (0,0) {Document $d$};
\node[anchor=north,align=center,font=\scriptsize] at (0.1,-0.75)
  {bank: source $d_1$,\\distractors $d_2,\dots,d_K$};
\foreach \x/\n/\t in {\xa/w0/Layer $0$,\xb/w1/Layer $1$,\xc/wl/Layer $l$}
  \node[frozen,minimum width=2.3cm,minimum height=0.8cm] (\n) at (\x,0) {\t};
\node at ($(w1)!0.5!(wl)$) {$\cdots$};
\node (wend) at (\xc+2.0,0) {$\cdots$};
\draw[flow] (d3.east |- doc) -- (w0);
\draw[flow] (w0) -- (w1);
\draw[flow] (w1.east) -- ++(0.6,0);
\draw[flow] ($(w1)!0.5!(wl)+(0.4,0)$) -- (wl);
\draw[flow] (wl) -- (wend);
\foreach \x/\n/\i in {\xa/h0/0,\xb/h1/1,\xc/hl/l}
  \node[trained,minimum width=2.3cm,minimum height=0.85cm] (\n) at (\x,-2.0)
    {Perceiver $\psi$};
\foreach \x/\w/\h/\i in {\xa/w0/h0/0,\xb/w1/h1/1,\xc/wl/hl/l}{
  \draw[tap] (\x-1.55,0) -- (\x-1.55,-1.0) -- (\x,-1.0) -- (\h.north);
  \node[anchor=west,font=\scriptsize] at (\x+0.05,-0.95) {layer-$\i$ input};
}
\foreach \x/\n/\i in {\xa/a0/0,\xb/a1/1,\xc/al/l}
  \node[gen,minimum width=2.3cm,minimum height=0.75cm] (\n) at (\x,-3.45)
    {LoRA};
\foreach \h/\a in {h0/a0,h1/a1,hl/al} \draw[tap] (\h) -- (\a);
\node[anchor=west,align=left,font=\scriptsize] at (\xc+1.35,-3.45)
  {For a bank, the factors of\\$d_1,\dots,d_K$ are concatenated\\along the rank dimension:\\$8K$ components, unit weight};
\node[rowlabel] at (-1.2,-4.55) {Serving time (every request)};
\node[input,minimum width=2.2cm,minimum height=0.9cm] (q) at (0,-5.7) {Question\\(no document text)};
\foreach \x/\n/\t in {\xa/s0/Layer $0$,\xb/s1/Layer $1$,\xc/sl/Layer $l$}{
  \node[frozen,minimum width=3.15cm,minimum height=2.0cm] (\n) at (\x,-6.0) {};
  \node[font=\footnotesize] at (\x,-5.2) {\t};
  \node[box,fill=white,font=\scriptsize,inner sep=2pt] (\n-attn) at (\x-.85,-5.75) {Attention};
  \node[box,fill=white,font=\scriptsize,inner sep=2pt] (\n-mlp) at (\x+.35,-5.75) {MLP\\up/gate};
  \node[box,fill=teal!8,draw=teal!70!black,font=\scriptsize,inner sep=2pt]
    (\n-down) at (\x+.35,-6.45) {Down-proj.};
  \node[draw=teal!70!black,circle,fill=white,inner sep=1pt,font=\scriptsize]
    (\n-add) at (\x+1.25,-6.45) {$+$};
  \draw[flow] (\n-attn) -- (\n-mlp);
  \draw[flow] (\n-mlp) -- (\n-down);
  \draw[flow] (\n-down) -- (\n-add);
}
\node (gap) at (9.55,-5.75) {$\cdots$};
\node (send) at (\xc+2.1,-6.45) {$\cdots$};
\node[box,minimum width=1.5cm,minimum height=0.9cm] (ans) at (\xc+3.6,-6.45) {Answer};
\draw[flow] (q) -- (s0-attn);
\draw[flow] (s0-add) -- ++(.5,0) |- (s1-attn);
\draw[flow] (s1-add) -- ++(.5,0) |- (gap);
\draw[flow] (gap) -- (sl-attn);
\draw[flow] (sl-add) -- (send);
\draw[flow] (send) -- (ans);
\foreach \x/\a/\s in {\xa/a0/s0,\xb/a1/s1,\xc/al/sl}
  \draw[tap] (\a.south) -- (\x,-4.75) -- (\x+1.25,-4.75) -- (\s-add.north);
\begin{scope}[shift={(-1.15,-7.65)},font=\scriptsize]
\node[input,minimum width=0.35cm,minimum height=0.25cm,inner sep=0pt] (k1) at (0,0) {};
\node[anchor=west] at (0.25,0) {input text};
\node[frozen,minimum width=0.35cm,minimum height=0.25cm,inner sep=0pt] at (2.0,0) {};
\node[anchor=west] at (2.25,0) {frozen model};
\node[trained,minimum width=0.35cm,minimum height=0.25cm,inner sep=0pt] at (4.3,0) {};
\node[anchor=west] at (4.55,0) {trained compiler};
\node[gen,minimum width=0.35cm,minimum height=0.25cm,inner sep=0pt] at (6.9,0) {};
\node[anchor=west] at (7.15,0) {generated adapter};
\end{scope}
\end{tikzpicture}}
\caption{Doc-to-LoRA. At write time the frozen model reads a document, and the
Perceiver compiler $\psi$ turns the input to each layer $l$ into rank-8 factors $A_{l,d}$ and
$B_{l,d}$. The retention probe reads the latent between the Perceiver and the output
head. One compiler serves all layers, with a separate output head for each layer. At serving time the prompt
holds only the question. The teal arrows show where LoRA adds
the document and shared bias contributions to the base down-projection output
$z$, producing the modified output $z'$ (Equation~\ref{eq:sum}).
The student-layer detail omits normalization and residual connections.
Each document in a bank is compiled
separately, and the factors of the source and the distractors are concatenated along
the rank dimension.}
\label{fig:architecture}
\end{figure*}
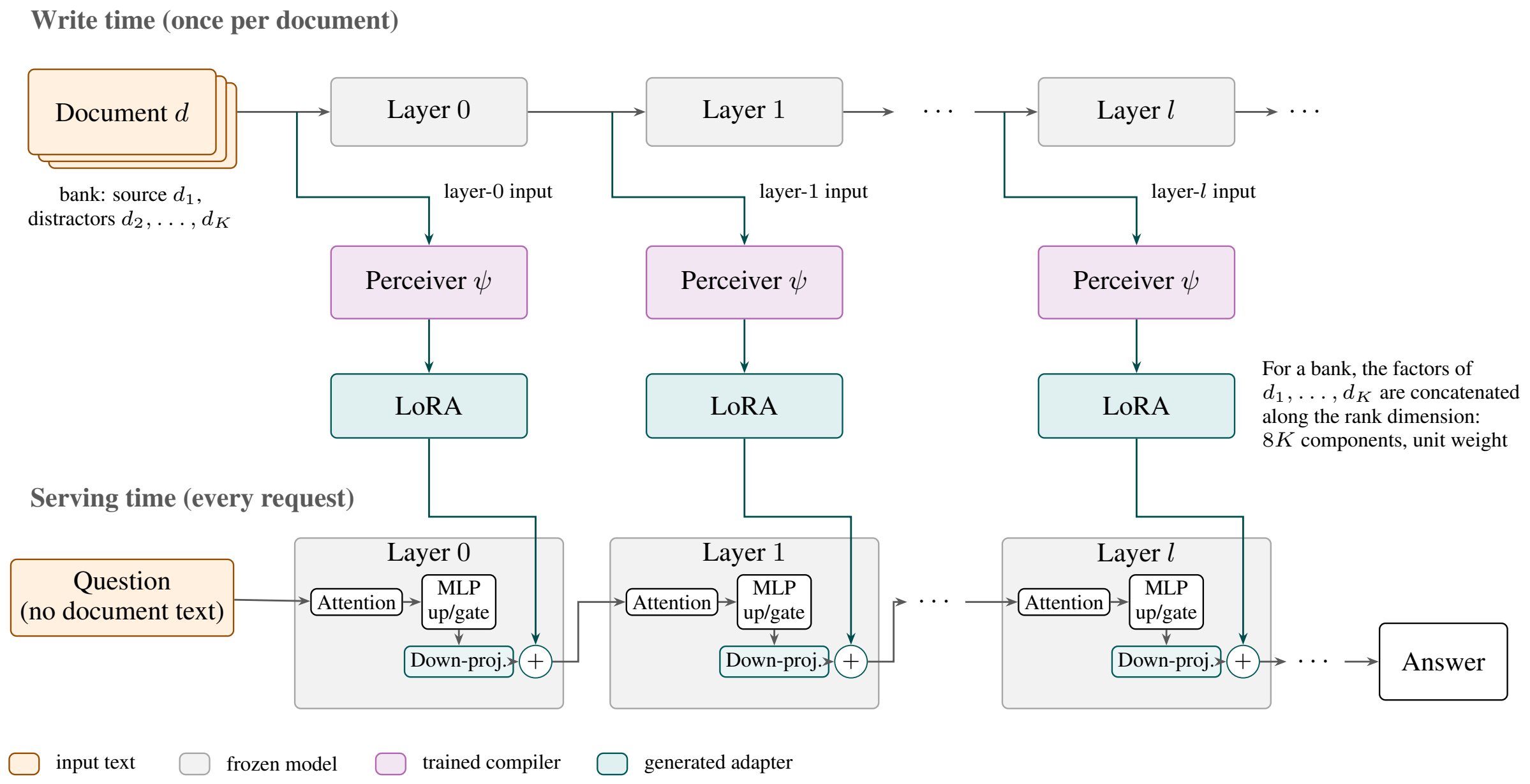

We use the released D2L compilers~\citep{charakorn2026doctolora} for
Gemma-2-2B-it~\citep{gemmateam2024gemma2} and Qwen3-4B-Instruct-2507~\citep{yang2025qwen3}.
The architecture, composition rule and bias adapter described here are from D2L.
We introduce the per-rank notation and gating interventions below.
Figure~\ref{fig:architecture} shows the native D2L pipeline. At write time, a
frozen copy of the target model reads a document $d$. A Perceiver-style
hypernetwork then converts the resulting layer activations into a LoRA
adapter: for each MLP layer $l$, it produces two rank-8 matrices,
$A_{l,d}$ and $B_{l,d}$, for the layer's down-projection. The hypernetwork was
trained by context distillation. The target model is frozen; compiler and
adapter updates are restricted to the fine-tuning and passage-fitting controls.
Thus, a document is compiled once
into model state and does not need to remain in the serving prompt.

We describe one layer at one token position, omitting layer and token indices
throughout the equations below. Let $x$ be the MLP down-projection input,
$z$ its base output, and $z'$ its modified output. The residual $h_{\mathrm{res}}$
enters the MLP branch after attention, and $h_{\mathrm{out}}$ is the block output
passed to the next layer. $\mathcal{N}$ is Gemma's post-MLP normalization,
or the identity for Qwen.

For document $d$, $A_d$ maps $x$ into rank space and $B_d$ maps back to the
down-projection output space. We include all adapter scaling in these factors,
absorbing the fixed implementation scale into $B_d$. The same convention
applies to the shared bias factors $A_0,B_0$. Native D2L sums over the supplied documents:
\begin{equation}
\begin{aligned}
z&=W_{\mathrm{down}}x\\
z'&=z+\sum_d B_dA_dx+B_0A_0x\\
h_{\mathrm{out}}&=h_{\mathrm{res}}+\mathcal{N}(z')
\end{aligned}
\label{eq:sum}
\end{equation}
Both document and bias adapters have rank $R=8$. The shared bias is added once;
the subscript 0 does not denote the first document or rank.
Appendix~\ref{sec:extended-methods} relates this matrix orientation to the
row-stored factors in the implementation.

The ICL reference gives the same model the same documents as text in its
context. We call this text-conditioned model the \emph{teacher}, because its
behavior is the reference that the compiled state is intended to reproduce. A
single-document comparison measures how closely the adapter reproduces the
model's behavior when it reads the document directly. For composition, a \emph{bank} is
the set of $K$ documents whose adapter contributions are summed for one request
(Figure~\ref{fig:architecture}). Each bank has one answer-bearing document,
called the \emph{source}, and $K-1$ documents that do not answer the question,
called \emph{distractors}. This arrangement tests whether the source's useful
information survives the addition of unrelated compiled states.
This is a distractor-robustness test, not a test of combining jointly necessary
sources. Source selection is an oracle control here; a general user-memory
system must also allow several memories to contribute to one request.

\subsection{Evaluation sets}
\label{sec:suites}

We use SQuAD, DROP, and ROPES, the factual-QA benchmarks used by the original
D2L study~\citep{rajpurkar2016squad,dua2019drop,lin2019ropes,charakorn2026doctolora}.
Unless a paragraph states otherwise, the metric is token F1. Qwen answers in full sentences, which
token F1 scores poorly against short reference spans, so for Qwen we also report
answer containment and compare each condition with Qwen's own single-document
score. Appendix~\ref{sec:set-details} gives the
construction and size of every set.

\begin{table}[t]
\centering
\caption{Factual-QA evaluation sets. Diagnostic baselines are specific to
the set and intervention runtime; their scores are not pooled.}
\label{tab:evaluation-sets}
\scriptsize\setlength{\tabcolsep}{3pt}
\begin{tabular}{@{}lrl@{}}
\toprule
Set & Questions & Purpose / results \\
\midrule
SQuAD test (200 sources) & 1,127 & Reader comparison, Fig.~\ref{fig:composition} \\
SQuAD development (180) & 1,260 & Localization, Fig.~\ref{fig:layers} \\
SQuAD diagnostic (57) & 400 & Readout, Table~\ref{tab:readout-summary} \\
Mixed-task diagnostic & 402 & Transplants/masks, Table~\ref{tab:causal} \\
SQuAD passage fits (160) & 864 & Rank/coverage, Appendix~\ref{sec:single-tables} \\
DROP/ROPES (612) & 1,934 & Localization, Fig.~\ref{fig:layers}(b) \\
\bottomrule
\end{tabular}
\end{table}

\subsection{Single-document diagnostics}
\label{sec:diagnostics}

\textbf{Retention (RQ1.1).} A trained probe predicts passage-token presence
from the compiler latent before conversion into LoRA factors. This measures
the latent, not the final adapter. Separately, we check whether incorrect D2L
answers match passage spans or another question's reference answer.

\textbf{Readout, compiler, and rank (RQ1.2).} We compare the post-block,
write-site, and gated readers defined in Section~\ref{sec:readers} with native D2L, fine-tune the compiler alone or
jointly with a reader, and fit rank-8--64 adapters to individual passages using
generated questions. We report passage-fit accuracy separately for answers
covered and not covered by those questions. We also test a question-conditioned compiler.

\textbf{Missing computation (RQ1.3).} Hidden-state transplants replace native
states with teacher states at question positions, question plus answer-history
positions, or all shared positions. Document-attention masks restrict which
positions can read the document.

\subsection{Multi-document diagnostics}
\label{sec:composition-diagnostics}

\textbf{Update size (RQ2.1).} Uniform normalization scales each document's
$A_d$ by $K^{-\beta}$. Development data select $\beta=0.5$ on both models.
This reduces update magnitude without selecting documents.

A \emph{source-only mask} retains the known source's adapter contribution and
removes all distractor contributions. A \emph{silent bank} removes all document
contributions. Both retain the shared bias. These diagnostic masks are prescribed;
Section~\ref{sec:readers} describes learned readers, including predicted rank gates.

\textbf{Damage and useful work by layer (RQ2.2).} A \emph{layer-prefix oracle}
retains native composition below layer $L$ and applies the source-only mask
from $L$ onward, measuring interference accumulated in the earlier layers.
A \emph{bank-silent schedule} instead silences the bank below $L$, measuring
the cost of withholding early source contributions. Additional interventions
delay masking or source application until the end of prompt processing,
testing whether later intervention can recover accuracy.
Appendix~\ref{sec:composition-details} specifies the schedules and cache handling.

\textbf{Separability (RQ2.3).} Membership probes predict source membership
from a layer's hidden state and a key for each rank. We measure held-out AUC
and test whether gates derived from the probes improve generated answers.

\subsection{Trained readers}
\label{sec:readers}

We test three readers that modify how frozen document adapters affect the model,
targeting single-document accuracy (RQ1.2) and distractor interference (RQ2.3).

In reader inputs, $A_d,B_d$ denote the factors for all documents in the bank.
A \emph{gated reader} predicts a vector $g_d$ of eight rank multipliers
for each document:
\begin{equation}
\begin{aligned}
g_d&=\boldsymbol{G_d}(h_{\mathrm{res}},x;A_d,B_d)\\
z'&=z+\sum_d B_d(\boldsymbol{g_d\odot} A_dx)+B_0A_0x
\end{aligned}
\label{eq:gated-reader}
\end{equation}
A \emph{write-site reader} uses the native down-projection output to predict
a correction, then adds it to that output:
\begin{equation}
z'\leftarrow z'+\boldsymbol{R_{\mathrm{write}}}(h_{\mathrm{res}},x,z';A_d,B_d)
\label{eq:write-site-reader}
\end{equation}
A \emph{post-block reader} instead adds its correction to the native block
output. It receives the individual rank effects and the shared bias effect:
\begin{equation}
\begin{aligned}
E_d&=B_d\operatorname{diag}(A_dx)\\
h'_{\mathrm{out}}&=h_{\mathrm{out}}+\boldsymbol{R_{\mathrm{post}}}(h_{\mathrm{out}};E_d,B_0A_0x)
\end{aligned}
\label{eq:post-block-reader}
\end{equation}
Here $\operatorname{diag}$ places a vector's entries on a diagonal matrix.
Each column of $E_d$ is one rank's contribution; the reader receives $E_d$
for every document.
This correction is added after normalization and residual addition.

The compiled factors and shared bias remain frozen. All readers are trained on
SQuAD; the single-document reference uses $K=1$ only, while the composition
readers use mixed-$K$ banks. Gates start at one and additive corrections at
zero, so every reader initially reproduces native D2L. Gated readers also
receive source-membership supervision. Appendix~\ref{sec:rank-gates} gives the
architectures and training protocols.

\subsection{D2L on MidBrain}
\label{sec:deployment}

\textbf{MidBrain on PersonaMem-v2 (RQ3).} The memories are produced by the
unchanged production MidBrain pipeline. It combines episodic and consolidated
retrieval, retaining episodic memories as unmodified source material.
Individual episodes are expected to be accessed infrequently, whereas the
smaller set of consolidated memories is reused more often. We therefore
compile consolidated memory, where repeated use can amortize compilation,
and retain retrieved episodes as text.

We use full history as text as the
baseline and compare MidBrain's
consolidated-memory prompt supplied as text with the same prompt compiled
jointly as one document per user. This includes persona/profile context and
consolidated entries. Retrieved episodes remain text, with the same episodes
supplied in both production conditions. We also compare these conditions
without episodic retrieval and test compiling consolidated entries separately.
For the matched multi-adapter comparison, we compile persona/profile context
as one document and each consolidated entry separately, retaining the same
retrieved episodes as text. We compare native summation with $K^{-0.5}$
normalization, where $K$ includes the persona/profile adapter.
We score the four option labels directly at the first answer position,
without generating chain-of-thought reasoning. This differs from the original
PersonaMem-v2 evaluation, which permits reasoning before the final
answer~\citep{jiang2025personamemv2}, and may contribute to differences in
absolute accuracy.

\textbf{Cost (RQ3).} We compare request-time prefill and decoding with
document text in context against prefill and decoding with its compiled
LoRA adapter. We calculate per-request MACs, attended positions, and
resident bytes for both conditions.
Separately, we measure latency, throughput, and GPU memory for ICL with full prefill and D2L on one
A100 using the released implementation. Throughput is measured with fixed
batch sizes of 1, 8, and 16, with one request and distinct memory per user;
this measures batched execution rather than concurrent-request scheduling.
Compilation is a write-time cost excluded from the request-time comparison.

\section{Results}
\label{sec:results}

\subsection{RQ1: one compiled document}
\label{sec:res-single}
\begin{figure}[t]
\centering
\begin{tikzpicture}
\begin{groupplot}[
 group style={group size=2 by 1,horizontal sep=0.75cm},
 scale only axis,width=0.34\columnwidth,height=2.7cm,
 xmode=log,log basis x=2,xtick={1,2,4,8,16},
 xticklabels={1,2,4,8,16},xmin=1,xmax=16,ymin=0,ymax=1,
 xlabel={Documents $K$},grid=major,
  tick label style={font=\scriptsize},label style={font=\scriptsize},
  title style={font=\scriptsize},
  legend style={font=\scriptsize,draw=none,legend columns=2}]
\nextgroupplot[title={Gemma-2-2B (F1)},ylabel={Token F1},legend to name=compositionlegend]
\addplot[black,mark=square*,thick] coordinates {(1,.872696)(2,.863)(4,.858)(8,.849)(16,.825)};
\addlegendentry{ICL}
\addplot[red!75!black,mark=*,thick] coordinates {(1,.755174)(2,.618)(4,.445)(8,.200)(16,.026)};
\addlegendentry{Native D2L}
\addplot[blue!75!black,mark=triangle*,thick] coordinates {(1,.755174)(2,.603575)(4,.421341)(8,.282533)(16,.228939)};
\addlegendentry{Normalized}
\addplot[teal!80!black,mark=diamond*,thick] coordinates {(1,.760839)(2,.624)(4,.465)(8,.268)(16,.117)};
\addlegendentry{Post-block reader}
\addplot[violet!80!black,mark=pentagon*,thick,dashed] coordinates {(1,.758)(2,.633)(4,.492)(8,.348)(16,.251)};
\addlegendentry{Write-site reader}
\addplot[orange!85!black,mark=x,thick,dashdotted] coordinates {(1,.750183)(2,.605258)(4,.417680)(8,.288555)(16,.184348)};
\addlegendentry{Gated reader}
\nextgroupplot[title={Qwen3-4B (containment)}]
\addplot[black,mark=square*,thick] coordinates {(1,.880213)(2,.876664)(4,.882875)(8,.891748)(16,.893523)};
\addplot[red!75!black,mark=*,thick] coordinates {(1,.683230)(2,.484472)(4,.321207)(8,.069210)(16,0)};
\addplot[blue!75!black,mark=triangle*,thick] coordinates {(1,.683230)(2,.433895)(4,.270630)(8,.197870)(16,.147294)};
\addplot[teal!80!black,mark=diamond*,thick] coordinates {(1,.633540)(2,.471162)(4,.342502)(8,.187223)(16,.072760)};
\addplot[orange!85!black,mark=x,thick,dashdotted] coordinates {(1,.619343)(2,.435670)(4,.271517)(8,.180124)(16,.104703)};
\end{groupplot}
\end{tikzpicture}
\par\smallskip\ref{compositionlegend}
\caption{Accuracy as the number of documents $K$ increases: token F1 for Gemma
and answer containment for Qwen. D2L falls below ICL already at $K=1$.
ICL stays relatively stable as distractors are added, while native D2L,
normalization, and the readers all degrade. All plotted readers are mixed-$K$.
Gemma includes all three readers; Qwen includes post-block and gated readers.
Gated readers use one seed, the other Gemma readers three.}
\label{fig:composition}
\end{figure}
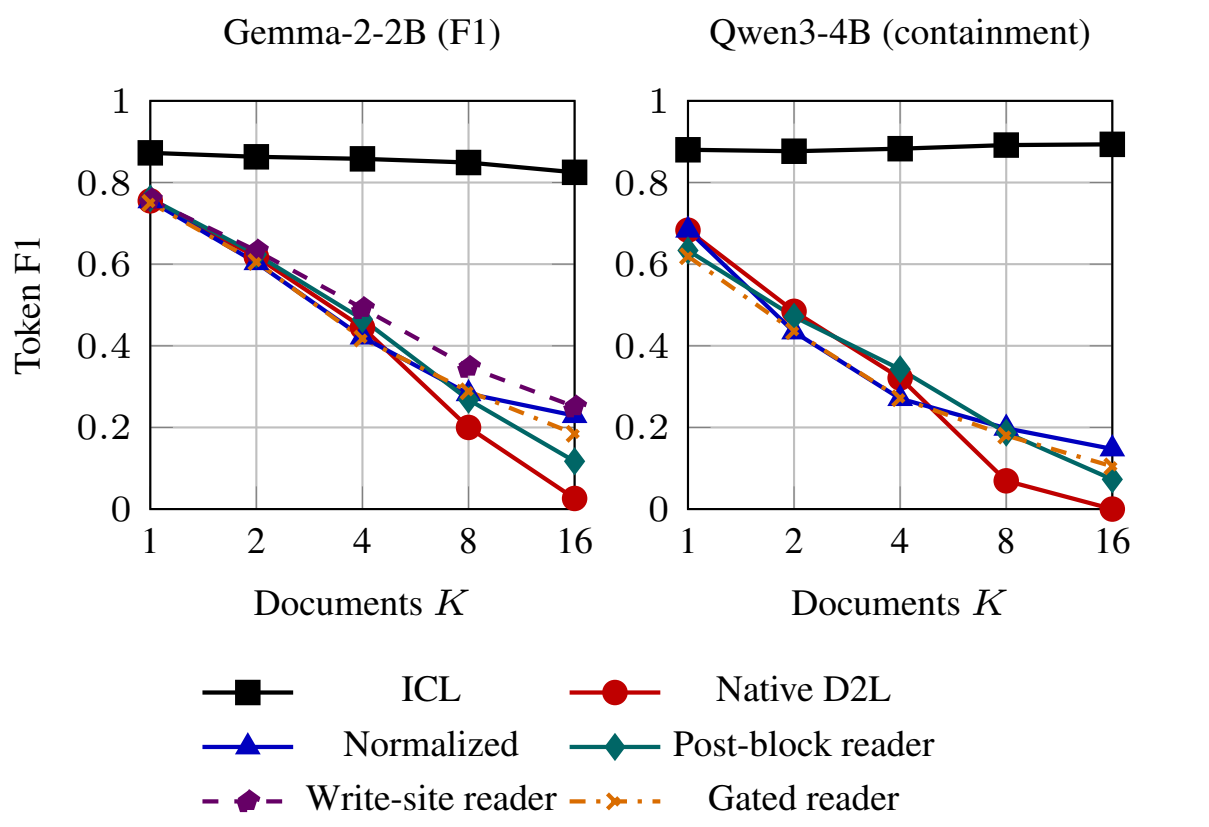

As can be seen in Figure~\ref{fig:composition}, D2L falls below ICL even at
$K=1$, and adding distractors widens the gap. The trained readers recover only
part of the loss and do not prevent accuracy from declining as more documents
are added. We first examine the single-document gap.

\subsubsection{Retention and attribution (RQ1.1)}
Using the answer-matching procedure described in Section~\ref{sec:diagnostics},
we compare incorrect responses with reference answers to other questions about
the same passage. Among teacher-correct questions each model misses, 46\% of
D2L's responses match another passage answer, compared with 10\% for the base
model. This describes passage-content attribution rather than a chance-corrected
binding error rate, since the base model has no document access.
The separate token-presence probe detects answer tokens in the
compiler latent even when generation fails, although it does not establish
whether those tokens survive conversion into LoRA factors.

\subsubsection{Readout and passage fitting (RQ1.2)}
As shown in Table~\ref{tab:readout-summary}, the $K=1$-only post-block readers
improve over native D2L but remain below ICL. Using individual rank effects
gives only a small improvement over summed effects or compiler latents.
Changing reader width or layer sharing does not close the remaining gap.
Compiler fine-tuning and
question-conditioned writes reduce training losses without improving
accuracy on held-out questions.

\begin{table}[t]
\centering
\caption{Post-block readers improve on native D2L but remain below ICL.
The three readers are trained at $K=1$ only on gold answers.}
\label{tab:readout-summary}
\small
\begin{tabular}{@{}lr@{}}
\toprule
Readout & F1 \\
\midrule
ICL (baseline) & .882 \\
Native D2L & .755 \\
Reader: summed adapter effect & .781 \\
Reader: compiler latent & .785 \\
Reader: individual rank effects & .789 \\
\bottomrule
\end{tabular}
\end{table}

Fitting adapters separately to each passage also leaves overall accuracy close
to native D2L on the 160-passage set. Answers covered by the generated training
questions approach teacher accuracy, while uncovered answers become less
accurate. At the selected budget of ten forward-KL steps, increasing rank
from 8 to 64 does not reliably improve overall accuracy or prevent the
decline on uncovered answers.
Appendix~\ref{sec:single-tables} provides supporting results for answer
attribution, the latent probe, and passage fitting.

\subsubsection{The document read at the answer positions (RQ1.3)}
As shown in Table~\ref{tab:causal}, replacing question-position states with
teacher states does not improve native D2L, whereas including answer-history
positions recovers part of the gap. Improving the question-position states
alone is therefore insufficient in this intervention.

The document-read masks in the same table show that both question and answer
positions contribute to ICL accuracy. Adding answer-position document reads
improves native D2L but does not reach the teacher, indicating that some of
the missing computation occurs during answer generation.

\begin{table}[t]
\centering
\caption{Effects of teacher-state transplants and document-attention masks
on answer accuracy. Gemma-2-2B, macro F1; each panel has its own measured baseline.}
\label{tab:causal}
\small
\begin{tabular}{@{}lr@{}}
\toprule
Intervention & F1 \\
\midrule
\multicolumn{2}{@{}l}{\emph{Hidden-state transplant}} \\
ICL (baseline) & .683 \\
Native D2L, self-transplant & .638 \\
Teacher states: question positions & .624 \\
Teacher states: question + answer history & .661 \\
\midrule
\multicolumn{2}{@{}l}{\emph{Direct document-read mask}} \\
ICL (baseline) & .688 \\
No positions read the document & .208 \\
Question positions only & .444 \\
Answer positions only & .427 \\
Native D2L & .633 \\
Native D2L + answer-position reads & .658 \\
\bottomrule
\end{tabular}
\end{table}

\subsection{RQ2: multi-document diagnostics}
\label{sec:res-composition}
\label{sec:res-gap}
\subsubsection{Composition and update size (RQ2.1)}
As can be seen in Figure~\ref{fig:composition}, adding distractors reduces
D2L accuracy substantially on both models, while ICL remains comparatively
stable. The normalization described in Section~\ref{sec:composition-diagnostics}
gives a modest improvement at larger bank sizes, while accuracy remains well
below ICL.

\subsubsection{Layer localization (RQ2.2)}
\label{sec:res-layers}
\begin{figure}[t]
\centering
\begin{minipage}[t]{.49\columnwidth}
\vspace{0pt}
\centering
\begin{tikzpicture}
\begin{axis}[
 width=\linewidth,height=3.4cm,grid=major,
 tick label style={font=\tiny},label style={font=\scriptsize},
 title={(a) Gemma/SQuAD},title style={font=\scriptsize},
 mark size=1pt,
 xlabel={Gate layer $L$},
 ylabel={F1},xmin=-1,xmax=23,ymin=0,ymax=0.95,xtick={0,4,8,13,18,22},
  legend style={font=\tiny,draw=none,at={(0.5,-0.42)},anchor=north,
  legend columns=1,cells={anchor=west}}]
\addplot[black,dashed,thick] coordinates {(-1,.870)(23,.870)};
\addlegendentry{ICL, $K=8$}
\addplot[red!75!black,dashed,thick] coordinates {(-1,.301)(23,.301)};
\addlegendentry{Normalized, $K=8$}
\addplot[blue!75!black,mark=*,thick] coordinates {(0,.740)(4,.708)(8,.611)(13,.466)(18,.388)(22,.234)};
\addlegendentry{Native below $L$, $K=8$}
\addplot[blue!40!black,mark=o,thick,densely dotted] coordinates {(0,.742)(4,.591)(8,.111)(13,.035)(18,.024)(22,.015)};
\addlegendentry{Native below $L$, $K=16$}
\addplot[teal!80!black,mark=square*,thick] coordinates {(0,.740)(4,.724)(6,.678)(8,.633)(13,.480)};
\addlegendentry{Silent below $L$, $K=8$}
\addplot[red!75!black,mark=triangle*,thick,only marks] coordinates {(0,.198)};
\addlegendentry{Native D2L, $K=8$}
\end{axis}
\end{tikzpicture}
\end{minipage}\hfill
\begin{minipage}[t]{.49\columnwidth}
\vspace{0pt}
\centering
\begin{tikzpicture}
\begin{axis}[width=\linewidth,height=3.4cm,grid=major,
 tick label style={font=\tiny},label style={font=\scriptsize},
 title={(b) Models and tasks},title style={font=\scriptsize},
 xlabel={Layer / depth},ylabel={F1 / native $K=1$},
 xmin=-0.03,xmax=0.9,ymin=0,ymax=1.05,
 xtick={0,.25,.5,.75},xticklabels={0,.25,.5,.75},
 legend style={font=\tiny,draw=none,at={(0.5,-0.42)},anchor=north,
 legend columns=1,cells={anchor=west}},
 mark size=1pt]
\addplot[blue!75!black,mark=*,thick] coordinates {(0,1)(0.154,.957)(0.308,.826)(0.5,.630)(0.692,.524)(0.846,.316)};
\addlegendentry{Gemma, SQuAD}
\addplot[blue!75!black,mark=*,thick,dashed,forget plot] coordinates {(0,1)(0.154,.978)(0.231,.916)(0.308,.855)(0.5,.649)};
\addplot[teal!80!black,mark=diamond*,thick] coordinates {(0,.990)(0.308,.797)(0.692,.462)};
\addlegendentry{Gemma, DROP/ROPES}
\addplot[teal!80!black,mark=diamond*,thick,dashed,forget plot] coordinates {(0,1)(0.231,.932)(0.308,.860)};
\addplot[violet!80!black,mark=triangle*,thick] coordinates {(0,1.004)(0.306,.686)(0.694,.155)};
\addlegendentry{Qwen, SQuAD}
\addplot[violet!80!black,mark=triangle*,thick,dashed,forget plot] coordinates {(0,1)(0.222,.958)(0.306,.905)};
\addlegendimage{black,thick}
\addlegendentry{Native below $L$}
\addlegendimage{black,thick,dashed}
\addlegendentry{Silent below $L$}
\end{axis}
\end{tikzpicture}
\end{minipage}
\caption{Layer sensitivity to distractors. (a) Absolute F1 on Gemma/SQuAD.
(b) Across models and tasks at $K=8$, normalized by model depth and native
$K=1$ accuracy. Distractors are removed from layer $L$ onward.}
\label{fig:layers}
\label{fig:layer-generality}
\end{figure}
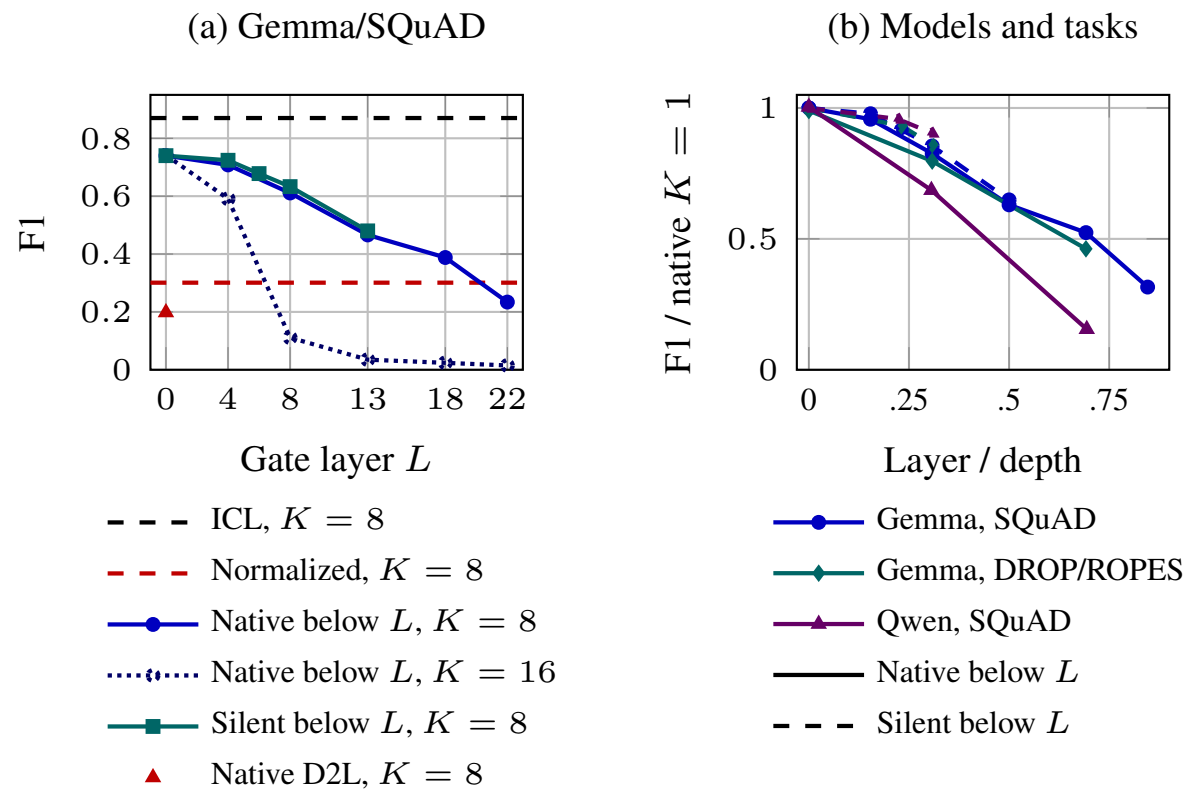

At $K=8$, applying the full bank in the initial layers gives similar measured
accuracy to withholding all document contributions there, with the source
alone applied in subsequent layers (Figure~\ref{fig:layers}(a)).
At layer 8 the two conditions reach .611 and .633 F1; at layer 13 they reach
.466 and .480. Interference accumulates across depth at this bank size,
whereas at $K=16$ retaining the full bank through layer 7 already reduces
F1 to .111, against .740 for the source alone.

Additionally, in Table~\ref{tab:prompt-interventions} we can see that delaying source application
until after prompt processing also reduces accuracy, despite using the known
source-only mask. Conversely, removing distractors only after prompt processing
does not recover the single-document result. The document contributions during
prompt processing therefore affect accuracy beyond the point at which the
adapters are changed.

\begin{table}[t]
\centering
\caption{Effects of delaying source application and distractor removal during
prompt processing. Gemma-2-2B, F1 at $K=8$. Cache replay updates cached
keys and values using source-induced residual corrections without fully
recomputing subsequent attention and MLP outputs.}
\label{tab:prompt-interventions}
\small
\begin{tabular}{@{}lr@{}}
\toprule
Condition & F1 \\
\midrule
ICL (baseline) & .870 \\
Native D2L, source alone & .740 \\
Native D2L, all documents & .198 \\
Uniform normalization & .301 \\
\midrule
\multicolumn{2}{@{}l}{\emph{Bank silent during prompt processing}} \\
Source at last prompt and response positions & .437 \\
Source applied with cache replay & .569 \\
\midrule
\multicolumn{2}{@{}l}{\emph{All documents active during prompt processing}} \\
Distractors removed at response positions & .313 \\
Also removed at last prompt position & .430 \\
\bottomrule
\end{tabular}
\end{table}

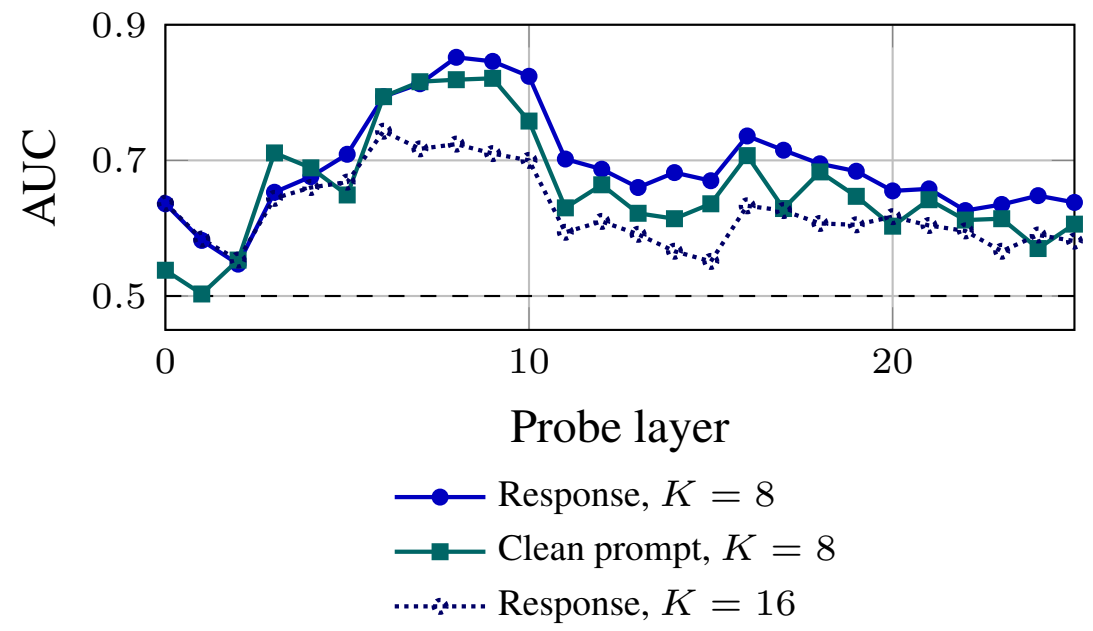
\begin{figure}[t]
\centering
\begin{tikzpicture}
\begin{axis}[width=.92\columnwidth,height=3.6cm,grid=major,
 tick label style={font=\scriptsize},label style={font=\small},
 xlabel={Probe layer},ylabel={AUC},
 xmin=0,xmax=25,ymin=.45,ymax=.9,
 xtick={0,10,20},ytick={.5,.7,.9},
 legend style={font=\scriptsize,draw=none,at={(0.5,-0.42)},anchor=north,
 legend columns=1,cells={anchor=west}}]
\addplot[black,dashed,thin,forget plot] coordinates {(0,.5)(25,.5)};
\addplot[blue!75!black,mark=*,mark size=1.2pt,thick] coordinates {
(0,.636)(1,.582)(2,.547)(3,.653)(4,.676)(5,.709)(6,.794)(7,.813)(8,.852)(9,.846)(10,.824)
(11,.702)(12,.687)(13,.660)(14,.682)(15,.670)(16,.736)(17,.715)(18,.695)(19,.684)(20,.655)
(21,.658)(22,.626)(23,.635)(24,.648)(25,.638)};
\addlegendentry{Response, $K=8$}
\addplot[teal!80!black,mark=square*,mark size=1.2pt,thick] coordinates {
(0,.538)(1,.503)(2,.553)(3,.711)(4,.689)(5,.649)(6,.794)(7,.816)(8,.819)(9,.821)(10,.758)
(11,.630)(12,.664)(13,.622)(14,.614)(15,.636)(16,.707)(17,.629)(18,.683)(19,.647)(20,.603)
(21,.642)(22,.612)(23,.614)(24,.570)(25,.606)};
\addlegendentry{Clean prompt, $K=8$}
\addplot[blue!40!black,mark=o,mark size=1.2pt,thick,densely dotted] coordinates {
(0,.637)(1,.584)(2,.553)(3,.644)(4,.660)(5,.668)(6,.743)(7,.717)(8,.724)(9,.710)(10,.700)
(11,.594)(12,.610)(13,.591)(14,.566)(15,.551)(16,.633)(17,.626)(18,.607)(19,.605)(20,.617)
(21,.604)(22,.596)(23,.566)(24,.590)(25,.581)};
\addlegendentry{Response, $K=16$}
\end{axis}
\end{tikzpicture}
\caption{Source-membership predictability by layer. Probe AUC on Gemma;
the dashed line marks chance.}
\label{fig:layer-probes}
\end{figure}

\label{sec:res-generality}
Figure~\ref{fig:layer-generality}(b) shows a similar pattern across models and
datasets: removing distractors earlier preserves more accuracy, while
withholding early source contributions also reduces accuracy.
However, the preregistered criterion that the loss at one-third depth reach
at least half the loss at two-thirds depth is not met: the observed fraction
is .36--.37 across all three model/task evaluations.

\subsubsection{Membership probes and gates (RQ2.3)}
\label{sec:res-probes}
Using the membership probes described in Section~\ref{sec:composition-diagnostics},
we measure how well each layer's hidden state and rank keys distinguish the
source from distractors. As shown in Figure~\ref{fig:layer-probes}, source membership is most readily
predicted around layers 6--10 on Gemma. Figure~\ref{fig:layers} shows that
document contributions in the preceding layers already affect answer accuracy,
so the strongest membership predictions become available after interference
has begun. The clean-prompt probe shows that source membership is also
detectable without first applying the document bank.

\subsubsection{Trained readers (RQ2.3)}
\label{sec:res-readers}
As shown in Figure~\ref{fig:composition}, none of the readers managed to learn
to reliably separate query-dependent facts from distractors.

On Qwen, the gated reader improves token F1 but reduces answer containment
at $K=1$. Its containment is close to normalization at $K=2,4,8$ and lower
at $K=16$, so the F1 gains do not establish better fact retention.

The localization results do not explain the reader failures by timing alone.
At $K=8$, a source-only oracle applied from layer 8 retains .611 F1 on the
development set, despite interference in the preceding layers. There is
therefore substantial scope for correction at intermediate depth; the tested
readers do not establish that this scope can be realized through their
architectures and training objectives.

\subsection{RQ3: D2L on MidBrain}
\label{sec:res-delivers}

\subsubsection{MidBrain on PersonaMem-v2}
\begin{table}[t]
\centering
\caption{MidBrain on PersonaMem-v2. Qwen3-4B accuracy and median memory
tokens per request. Adapter conditions retain the same retrieved episodes as text.}
\label{tab:user-memory-summary}
\small\setlength{\tabcolsep}{3pt}
\begin{tabular}{@{}lrr@{}}
\toprule
Memory representation & Accuracy & Tokens \\
\midrule
Full history as text (baseline) & .364 & 31{,}847 \\
\midrule
Episodic (text) & .376 & 8{,}736 \\
Consolidated (text) & .353 & 1{,}451 \\
Episodic + consolidated (text) & .390 & 10{,}159 \\
Episodic + consolidated (LoRA) & .391 & 8{,}736 \\
Episodic + consolidated (LoRA sum) & .274 & 8{,}736 \\
Episodic + consolidated (LoRA norm.) & .387 & 8{,}736 \\
\bottomrule
\end{tabular}
\end{table}

As shown in Table~\ref{tab:user-memory-summary}, compiling consolidated memory
jointly preserves similar measured accuracy to supplying it as text.
Summing separately compiled adapters substantially reduces accuracy even
with episodic text available. Normalization recovers most of this loss,
bringing accuracy close to joint compilation and text memory.
It should be noted that absolute scores are not directly comparable with the original PersonaMem-v2
evaluation because our protocol omits generated reasoning.

\subsubsection{Serving cost}
\begin{figure}[t]
\centering
\begin{tikzpicture}
\begin{groupplot}[group style={group size=2 by 1,horizontal sep=1.2cm},
 scale only axis,width=.31\columnwidth,height=2.6cm,grid=major,
 tick label style={font=\scriptsize},label style={font=\scriptsize},
 title style={font=\scriptsize},
 legend style={font=\scriptsize,draw=none,at={(.5,-.28)},anchor=north,legend columns=1}]
\nextgroupplot[title={(a) Analytical compute},
 xlabel={Memory tokens},ylabel={GMAC/request (log)},
 xmode=log,log basis x=2,ymode=log,
 xtick={150,1024,4096},xticklabels={150,1024,4096},
 ymin=100,ymax=20000]
\addplot[black,dashed,mark=square*,thick] coordinates {(150,583)(512,1544.8)(1024,2929.1)(2048,5781.3)(4096,11820.7)};
\addlegendentry{ICL: full prefill}
\addplot[teal!80!black,mark=diamond*,thick] coordinates {(150,188.8)(512,188.8)(1024,188.8)(2048,188.8)(4096,188.8)};
\addlegendentry{Native D2L}
\nextgroupplot[title={(b) Batched serving throughput},
 xlabel={Batch size},ylabel={Aggregate tokens/s},
 xtick={1,8,16},xmin=0,xmax=17,ymin=0,ymax=350]
\addplot[teal!80!black,dashed,mark=diamond*,thick] coordinates {(1,24.3)(8,165.3)(16,290.9)};
\addlegendentry{Gemma ICL}
\addplot[teal!80!black,mark=diamond*,thick] coordinates {(1,20.1)(8,157.8)(16,315.8)};
\addlegendentry{Gemma D2L}
\addplot[blue!75!black,dashed,mark=triangle*,thick] coordinates {(1,21.5)(8,122.0)(16,162.5)};
\addlegendentry{Qwen ICL}
\addplot[blue!75!black,mark=triangle*,thick] coordinates {(1,17.1)(8,134.9)(16,269.4)};
\addlegendentry{Qwen D2L}
\end{groupplot}
\end{tikzpicture}
\caption{Recurring compute and serving throughput. (a) Gemma, one document,
40 question tokens and 32 output tokens. (b) One A100, four documents per user
and 64 output tokens per user. Each batch contains one request per user,
with distinct user-specific memory. Throughput includes prefill and decoding;
ICL performs full prefill per request. Compilation is excluded.}
\label{fig:serving-tradeoff}
\end{figure}
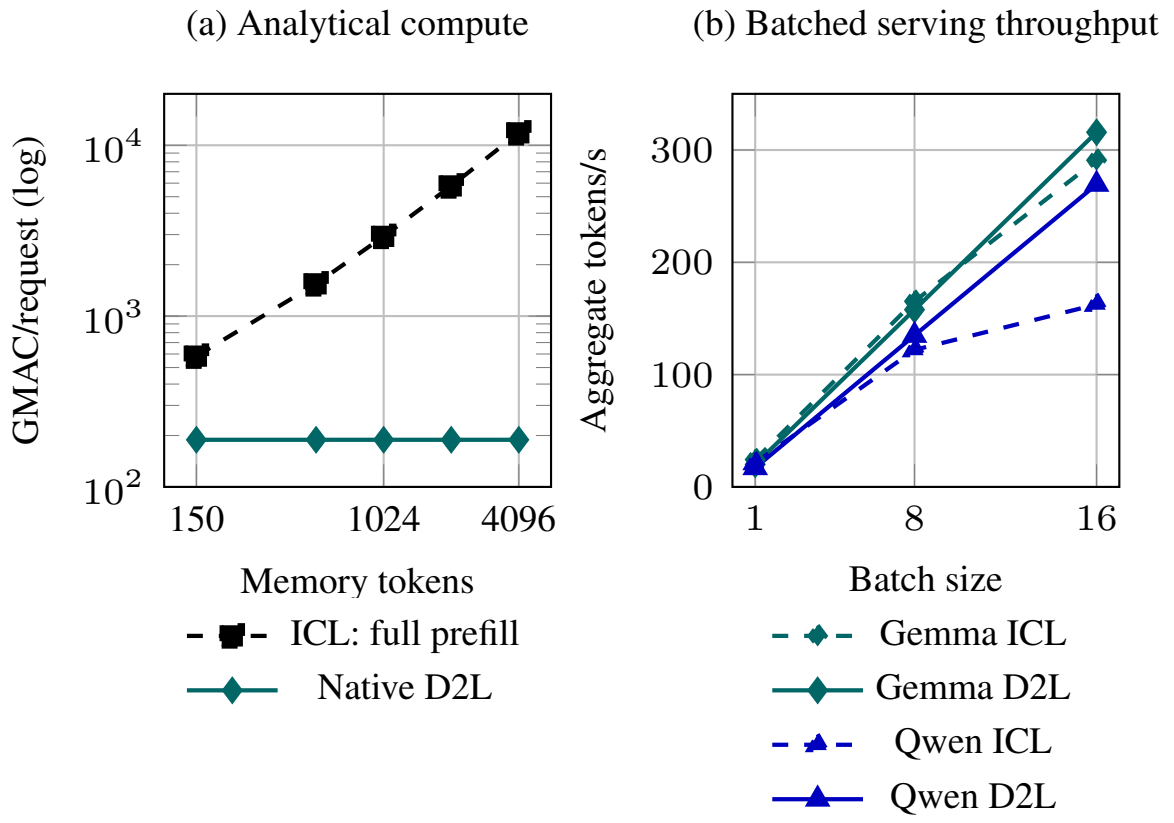

As shown in Figure~\ref{fig:serving-tradeoff}(a), D2L's recurring compute
advantage grows with document length: ICL processes more document tokens,
whereas the compiled adapter has fixed size. These estimates cover
request-time prefill and decoding, excluding compilation.

As shown in Figure~\ref{fig:serving-tradeoff}(b), increasing batch size from
1 to 16 raises D2L's throughput relative to ICL from $0.83\times$ to
$1.09\times$ on Gemma and from $0.80\times$ to $1.66\times$ on Qwen.
We hypothesize that batching amortizes model-weight reads and execution
overhead, making the cost of ICL's longer contexts more prominent.
At small batch sizes, the additional LoRA operations may instead cost more
time than the document computation they replace, particularly when execution
is limited by memory access or kernel-launch overhead rather than arithmetic.

\section{Discussion}
\label{sec:discussion}

Our results identify a gap between retaining document information and using it
reliably. Incorrect D2L responses frequently match answers to other questions
about the same passage, while passage-token information remains detectable in
the compiler latent. Passage-specific fitting improves answers covered by its
supervision but degrades uncovered answers, suggesting that internalization
must preserve question--answer associations as well as content. Increasing
adapter rank alone did not resolve this problem at the tested fitting budget.

The distribution of composition damage depends on bank size. On Gemma, with eight
documents, interference accumulates across depth, and applying the full bank
in the initial layers provides little advantage over withholding document
contributions there. With sixteen documents, much of the damage has already
occurred by layer 8. Although this constrains late interventions, the
source-only oracle at $K=8$ retains substantial accuracy at intermediate depth,
so timing alone cannot account for the trained readers' failures.
The results identify a limitation of the tested composition and reader
configurations, while leaving open whether more effective selectivity can
be learned during compilation or reading.

Although compiling consolidated memory largely preserves its benefit on
PersonaMem, the resulting token savings are modest because episodic memories
remain text, and the serving measurements show that lower analytical compute
does not consistently translate into higher throughput. These limited gains
must be weighed against compilation cost and the attribution and interference
failures observed in factual QA, leaving little evidence that D2L offers a
compelling alternative to text-based memory in the production settings studied
here.

\subsection{Limitations and threats to validity}
We test only two models and one type of trained hypernetwork. It is unclear
whether other hypernetwork architectures or training methods would produce
the same failures.

\subsection{Future work: efficient online and continual learning}
At MidBrain, we want to turn semantic and procedural memories into internalized
behavior, so a model does not have to spend as much compute reading the same
memories in context on every request. D2L provides one way to do this, but the
failures studied here motivate continuing with other continual-learning
approaches.

We are particularly interested in learning from interactions in which a memory
is actually used. Such interactions provide both the request and an example
of how the memory should affect the response, offering supervision beyond the
memory text alone. For procedural memory, the objective is to learn how to
carry out a preference or instruction and recognize when it applies. Learning
across several interactions may help preserve this relationship while reducing
the need to repeatedly supply the instruction as text.

We believe online-learning may provide an interesting avenue of research.
Learning personalized behaviors at inference time could reduce the need for
separate training infrastructure. In principle, much of the information needed
for learning is already available during inference; if the necessary
activations are retained, what remains is largely a backward pass and parameter
update. However, online-learning also introduces challenges, including
catastrophic forgetting and difficult convergence properties.

Over time, these updates must accommodate new information without disrupting
behavior that remains valid. Our interest is therefore in bounded per-user
state that can accumulate compatible memories, adapt when preferences change,
and preserve the applicability of earlier instructions. We want to evaluate
this on sequences of interactions, measuring whether learning improves later
responses, how well previously acquired behavior is retained, and how much
context and serving computation can be avoided. This would connect the
immediate cost of an update to its value over the subsequent lifetime of the
memory.

\bibliography{bib/references}
\bibliographystyle{icml2026}

\clearpage
\appendix
\section{Evaluation and implementation details}
\label{sec:extended-methods}

\subsection{Implementation convention}
For the interventions in Section~\ref{sec:d2l}, we map the released
row-stored factors to our notation as
$A_{l,d}=\widetilde A_l(d)$ and $B_{l,d}=s\widetilde B_l(d)^\top$.
The fixed implementation scale is $s=2R^{3/2}$ with $R=8$;
learned scalers are already included in the stored factors.
The same convention applies to the shared bias adapter.

\subsection{Set construction}
\label{sec:set-details}
All factual-QA evaluations use greedy generation with a 32-token limit, matched
prompts, and D2L's evaluation normalization.
Scores from different sets are never pooled, because the sets answer different
diagnostic questions and have different document construction histories.

The main comparisons use held-out development and test sets. Configurations
are selected on development data. The same test set was used sequentially for
the post-block, selected write-site, and gated-reader evaluations. The
preregistered rule required an improvement over native D2L at $K=1$ and over
normalization at $K=4$ and $K=8$. A post-block reader trained only at $K=1$
provides the single-document reference.

\textbf{Test set.} The test set holds 200 SQuAD validation source
paragraphs with 1,127 questions, plus 40 development sources with 200 questions from
disjoint articles. From 2,067 SQuAD validation paragraphs, 842 referenced by earlier
experiments through document ID, content hash, question index, or 10-word overlap are
excluded, as are paragraphs of the first 1,000 validation questions used by D2L for
model selection. Articles are partitioned between development and test. Up to eight
paragraphs per article with at least five questions serve as sources; remaining
paragraphs serve as distractors. Random banks are nested across $K$ with distractors
from other articles in the same split and a fixed random source position, so a nested
bank reuses the same source and adds documents as $K$ grows.
Bank identifiers and source positions are persisted.

\textbf{Development set.} This set contains the first 180 unique SQuAD-train
contexts in dataset order and their 1,260 questions. The released compiler's training
definition holds these contexts out. Nested random banks with
$K\in\{1,2,4,8,16\}$ draw distractors from the same 180-document pool and use a
hashed source position. These contexts had served as fresh
validation data for earlier single-document readers and were excluded from every
reader's training rows, validation rows, and distractor pools in this paper.
Development data from this set determine settings such as the normalization exponent
and reader arm. We evaluate the selected configurations on the test set.

\textbf{Diagnostic sets.} The first is a 57-document, 400-question SQuAD set for
readout and retention diagnostics. The second is a 402-question SQuAD, DROP, and
ROPES validation set for transplants and masks. The third is a previously unused
160-passage, 864-question SQuAD set for per-passage fits. Its passages lie in the
compiler's meta-training corpus. These sets locate the source of a failure or establish a
ceiling for a particular intervention. We do not pool their scores across tasks.

\textbf{DROP/ROPES set.} After excluding D2L's selection rows, this set contains
612 test sources with 1,934 questions: 527 DROP documents with 1,530 questions,
capped at three per document, and 85 ROPES documents with 404 questions. The
94 DROP documents never used by any earlier experiment are flagged separately.

\subsection{Single-document protocols (RQ1)}
These protocols support Section~\ref{sec:res-single} and
Tables~\ref{tab:readout-summary} and~\ref{tab:causal}.
\textbf{Readout comparisons.} Shared and independent width-32 post-block corrections
and write-site readers of width 32 and 2,304 use forward KL against the ICL teacher.
The representation comparison holds a width-32 post-block reader and gold-answer
cross-entropy fixed while varying its memory input: summed adapter effect,
individual rank effects, or compiler latents. Each uses 52,354 SQuAD training
questions, 4,365 updates, 400 document-disjoint validation questions, and 400
held-out questions from 57 documents. Compiler-only and joint ablations train the
hypernetwork alone or with the reader under the same schedule and select by fresh
validation F1 with checkpoint zero eligible. \textbf{Retention and attribution.} A
kernel-ridge probe predicts passage-token presence from the frozen
$26\times8\times512$ pre-head latent, trained on 2,000 separate documents.
Incorrect responses on teacher-correct questions (F1 $\geq0.5$) are matched to
passage spans and other questions' reference answers. \textbf{Transplants and
masks.} On the 402-question set, native hidden states are replaced by ICL states at
selected position sets at every layer boundary, with cache-free replay and teacher
states recomputed on the same answer history. Self-transplants reproduce native
logits, and all-position transplants reproduce teacher logits. Masks preserve teacher prompt
geometry and restrict document reads by position set, with baselines re-measured in
the same runtime. \textbf{Per-passage fits.} Rank-8/16/32/64 down-projection
adapters are fitted on 160 passages with 864 natural questions by full-batch forward
KL on generated factual probes, 10 steps at learning rate $10^{-5}$ selected on 24
development passages. Strict probes use reference spans and exclude generated
questions with more than 0.5 token-F1 overlap with natural questions.
Natural questions are stratified by whether the probes elicited their answer.
\textbf{Question-conditioned writes.} A shared residual encoder reads the
compiler latents and prompt-position states and uses the frozen LoRA head. Variants
train at $K=1$ with 159,040 questions under teacher KL or an additional
same-passage wrong-question hinge.

\subsection{Layer interventions (RQ2)}
\label{sec:composition-details}
These interventions support Figure~\ref{fig:layers}
and Table~\ref{tab:prompt-interventions}.
\textbf{Oracles and schedules.} Layer-prefix oracles start the source-only mask at
$L\in\{0,4,8,13,18,22\}$ of Gemma's 26 layers at $K=8$ and $K=16$. Bank-silent
schedules use $L\in\{0,4,6,8,13\}$. Cross-model and cross-task evaluations use gate layers 0, 8, 18 and
silent layers 6, 8 on Gemma-2-2B with the DROP/ROPES set, and gate layers 0, 11, 25 and
silent layers 8, 11, the same fractions of its 36 layers, on Qwen3-4B with the
development set. Each evaluation has its own ICL, native, normalized, and single-document
rows. All gated conditions run inside the $K$-bank's
compiled batch with the correction of Equation~\ref{eq:gate}. The layer-0 oracle
reproduces native $K=1$'s response string on all but 24--30 of 1,260 development-set
questions on Gemma, 71 of 1,934 on DROP and ROPES, and 115 of 1,260 on Qwen3-4B,
whose responses average 13 words. The remaining strings diverge at greedy ties, and
the paired F1 contrasts against native $K=1$ lie within $\pm0.004$ with intervals
including zero. Schedule A keeps $g=0$ at every prompt position but the
last, where it applies the mask, and applies the mask at every decode position and
layer in one causal pass. Schedule C performs a clean prefill of the prompt, then for
each layer computes the change after post-MLP normalization,
\begin{equation}
\Delta h_{\mathrm{replay}}
=\mathcal{N}(z'_{\mathrm{masked}})-\mathcal{N}(z'_{\mathrm{silent}})
\label{eq:cache-replay}
\end{equation}
where the silent output retains the shared bias and the masked output adds
only the source's document contribution. It shifts the residual entering
every later layer by the cumulative change, recomputes that layer's keys and values
from the shifted residual with rotary embeddings, and overwrites the cache positions.
Later layers' attention and MLP outputs are not re-evaluated. The last prompt
position is then recomputed exactly under the masked bank over the replayed cache
and decoding runs with the masked bank. A manual greedy loop over the static hybrid
cache reproduces the library's generation, and the replay is exact at the first
affected layer. Both properties are unit-tested. Response-only oracles keep native composition
at every prompt position and layer and apply the mask at every layer of the generated
positions, optionally also at the last prompt position. The layer-6 source-only
reference uses the same banks.

\subsection{Probe and reader implementation (RQ2)}
\label{sec:rank-gates}
The following definitions specify the probes and readers in
Sections~\ref{sec:composition-diagnostics} and~\ref{sec:readers}.
Let $a_{k,r}^{\top}$ and $b_{k,r}^{\top}$ be row $r$ of the implementation's
$\widetilde A$ and $\widetilde B$ factors for document
$d_k$. Expanding the native update into its rank contributions gives
\begin{equation}
\begin{aligned}
c_{k,r}&=a_{k,r}^{\top}x\\
z'&=z+s\sum_{k=1}^{K}\sum_{r=1}^{R}c_{k,r}b_{k,r}+B_0A_0x
\end{aligned}
\label{eq:rank-bank}
\end{equation}
Each row of $\widetilde A$ computes a coefficient from $x$, which multiplies the
corresponding row vector from $\widetilde B$, transposed. Factors are fixed after compilation;
coefficients depend on the current activation.

\textbf{Probes.} The signed strength of a document rank and its two features are
\begin{equation}
\begin{aligned}
m_{k,r}&=s\,c_{k,r}\lVert b_{k,r}\rVert_2\\
\phi(m)&=[m,\ \operatorname{sign}(m)\log(1+|m|)]^{\top}
\end{aligned}
\label{eq:strength-features}
\end{equation}
The membership probe combines a state--rank score with these strength features:
\begin{equation}
\ell_{k,r}=
\frac{\mathbf q(h)^{\top}\mathbf k_{k,r}}{\sqrt{64}}
+w^{\top}\phi(m_{k,r})+b
\label{eq:probe-score}
\end{equation}
Here $h$ is the captured MLP-input residual, distinct from the down-projection
input $x$. The query is $\mathbf q(h)=W_q^{\top}\mathrm{LN}(h)$,
and $\mathbf k_{k,r}$ is the rank key after a learned projection. The two
terms measure state--rank agreement and signed strength, respectively;
$w$ weights the strength features and $b$ is a learned score bias.
We fit the scorer with one AdamW pass at learning rate $10^{-3}$ over 1,500
$K$-balanced banks ($K\in\{2,4,8,16\}$) from the reader-training rows. The objective
is binary cross-entropy on rank membership. We score every (question, random bank)
pair in the development set at $K=8$ and $K=16$ for response-position
probes and at $K=8$ for the clean-prompt probe.
For the reported document-level AUC, we average
the eight rank logits of each document. For per-question AUC, we first average over
the query positions. For
clean-prefill collection the bank is cancelled during the forward pass while the
coefficients and strengths are still computed from the bank.

\textbf{Reader training.} Readers are trained for 4,291 updates on SQuAD training
questions with the development-set documents excluded. All readers use AdamW with
learning rate $10^{-4}$, weight decay $10^{-4}$ and gradient clipping at 1.
Each update uses 12 questions, with one question per backward pass and data-parallel
training across eight A100 GPUs. Mixed-$K$ vector-correction readers use balanced
$K\in\{1,2,4,8,16\}$ blocks and cached frozen
compiler latents. We validate every 500 updates by greedily generating answers to
200 fresh questions (40 per $K$). We keep the best checkpoint by fresh macro-F1
and the last checkpoint. The post-block reader uses gold-answer cross-entropy with the
tokenizer's end-of-sequence token appended. The write-site and gated readers use the
model's chat terminator, after finding that the appended token was a separate stop
symbol on Gemma and that an answer-only target removed stop supervision. Write-site
readers compute keys from the layer-normed $B$ rows and a scaled coefficient, values
from the raw rank effects, add a sum-pooled bank feature, and emit the correction in
the MLP branch's units. The deep variant adds slot embeddings from $(a_{k,r},b_{k,r})$ through
two transformer blocks over the $8K$ slots, computed once per bank, and two
cross-attention blocks from four read tokens. Gated readers use the strength
features of Equation~\ref{eq:strength-features}, but add their projection to
the rank key before taking the query--key product. The latent-assisted variant
also receives compiler latents:
\begin{equation}
\ell_{k,r}=
\frac{\mathbf q(u)^{\top}
 [\mathbf k_{k,r}+W\phi(m_{k,r})]}{\sqrt{32}}
+b+4
\label{eq:trained-gate-score}
\end{equation}
Here $u$ is the layer-normalized read state produced by the reader, not the
probe's captured residual $h$. The gate has its own query projection $\mathbf q$,
strength projection $W$, and score bias $b$, distinct from the probe parameters.
The gate is
\begin{equation}
g_{k,r}=\frac{\sigma(\ell_{k,r})}{\sigma(4)}
\label{eq:trained-gate}
\end{equation}
The query projection and score bias start at zero, so $\ell_{k,r}=4$ and
$g_{k,r}=1$ at initialization. The upper bound is
$1/\sigma(4)\approx1.0183$. The membership loss is the binary cross-entropy of $\sigma(\bar\ell_k)$, the
document-mean logit, against bank membership, averaged over documents, layers, and
response positions, with weight 1. Selection used the macro over $K\in\{1,4,8\}$ of
the variant's paired difference from the reference (native at $K=1$,
normalization at $K\geq4$). Ties were assigned to the cheaper arm. The test-set evaluation
reused the recorded ICL, native, and normalized rows. Before evaluation,
native outputs were checked against 12 recorded examples.

\textbf{Applying the gates.} Each gate scales one document rank:
\begin{equation}
z'_{\mathrm{gated}}=z+s\sum_{k,r}g_{k,r}c_{k,r}b_{k,r}+B_0A_0x
\label{eq:gated-ranks}
\end{equation}
The native adapter has already been applied when the gating hook runs, so the
implementation adds only the difference:
\begin{equation}
z'_{\mathrm{gated}}-z'_{\mathrm{native}}
=s\sum_{k,r}(g_{k,r}-1)c_{k,r}b_{k,r}
\label{eq:gate}
\end{equation}
The base projection and shared bias cancel. This correction is added before
post-MLP normalization. Gates of 1 leave the native output unchanged; gates of
0 remove all document contributions while keeping the bias.

\subsection{MidBrain and serving protocols (RQ3)}
These protocols support Table~\ref{tab:user-memory-summary} and
Figure~\ref{fig:serving-tradeoff}.
\textbf{PersonaMem.} The evaluation uses 195 test personas and 4,878 questions. The
fixed memory snapshot contains consolidated entries and
persona/profile context derived from each history with no benchmark questions. Five
development personas are excluded. The provider returns 100 episodes per question,
retained across paired production conditions. Memory formatting follows the
production configuration; evaluation uses an option-label-only answer prompt.
The answering model is Qwen3-4B-Instruct-2507 with its released D2L
checkpoint. Joint compilation includes persona, profile, and all entries in one
document. Per-entry compilation uses one adapter per entry plus one for
persona/profile.
Label-only scoring selects the most likely option label in one
forward pass. Paired persona-cluster intervals use 4,000 bootstrap resamples.
\textbf{Cost.} For a memory of $M$ tokens in $K$ documents and a request with $Q$
input and $T$ output tokens, write $N=Q+T$. With $P$ the linear MACs per token, $a$ the attention
MACs per attended position, and $\rho$ the adapter MACs per token and document,
\begin{align}
C_{\mathrm{ICL}} &\approx (M+N)P+\tfrac{a}{2}(M+N)^2\label{eq:cost-icl}\\
C_{\mathrm{D2L}} &\approx N\big(P+\rho(K+1)\big)+\tfrac{a}{2}N^2\label{eq:cost-d2l}
\end{align}
The $K+1$ adapters comprise $K$ documents and one shared bias. For storage,
let $p$ be parameters per adapter, $b$ bytes per parameter, $L$ layers,
$H$ KV heads, and $D$ the head dimension:
\begin{align}
S_{\mathrm{user}}&=Kpb\label{eq:storage-adapter}\\
S_{\mathrm{KV}}&=2LHDMb\label{eq:storage-kv}
\end{align}
The factor 2 counts keys and values. The shared bias adds $pb$ once to
model state, not to each user's stored document adapters. It is included in
per-request MACs. Measurements use one A100 (40\,GB), PyTorch 2.9.0+cu130, upstream D2L LoRA
execution, 64 forced output tokens, the median of five repetitions, and multi-user
batches of 1, 8, and 16 distinct four-passage memories.
Compilation is a write-time cost and is reported separately.
Factual-QA intervals are paired source-document bootstrap intervals with
10,000 resamples after averaging within each source. PersonaMem uses the
4,000-resample persona-cluster procedure above. Tables identify their averaging unit.

\section{Details supporting RQ1}
\label{sec:single-tables}
\subsection{Readout and attribution}
Table~\ref{tab:reader-ablations} provides the width, layer-sharing, and
compiler-training comparisons discussed in Section~\ref{sec:res-single}.
The representation comparison is reported in Table~\ref{tab:readout-summary}.

\begin{table}[t]
\centering
\caption{Single-document controls (Gemma; 400 questions from 57 documents).
Reader controls use teacher KL; compiler controls use gold-answer
cross-entropy and select checkpoint zero. Native F1 is .755; ICL is .882.}
\label{tab:reader-ablations}
\small\setlength{\tabcolsep}{3pt}
\begin{tabular}{@{}lr@{}}
\toprule
Condition & F1 \\
\midrule
Post-block, shared width 32 & .740 \\
Post-block, independent width 32 per layer & .744 \\
Write-site, width 32 & .751 \\
Write-site, width 2,304 & .735 \\
Compiler fine-tuning & .755 \\
Joint compiler and reader fine-tuning & .755 \\
\bottomrule
\end{tabular}
\end{table}

\begin{table}[t]
\centering
\caption{Attribution of incorrect responses on teacher-correct questions
in the 400-question diagnostic set (Section~\ref{sec:res-single}).
$n$ is the number of errors for each model; correctness uses F1 $\geq .5$.
The two matching categories overlap.}
\label{tab:fact-attribution}
\small\setlength{\tabcolsep}{3pt}
\begin{tabular}{@{}lrrr@{}}
\toprule
Model & $n$ & Passage span & Other answer \\
\midrule
Native D2L & 65 & 57\% & 46\% \\
Rank-effect reader & 59 & 63\% & 58\% \\
Base model & 291 & 11\% & 10\% \\
\bottomrule
\end{tabular}
\end{table}

Among teacher-correct questions with answer tokens in the probe vocabulary,
the mean answer-token percentile in the compiler-latent probe is .880 for
native-correct responses ($n=222$) and .849 for native-incorrect responses
($n=44$). These are the probe-covered subsets of 302 successes and 65 failures;
percentiles measure token scores within the probe vocabulary, not generation
accuracy or retention in the final adapter.

\subsection{Passage fitting and question conditioning}
\label{sec:res-refinement}
These tables give the supporting measurements for the passage-fitting and
question-conditioning controls in Section~\ref{sec:res-single}.

\begin{table}[t]
\centering
\caption{Strict answer-aware passage fits: learning rate $10^{-5}$, step 10
(160 documents, 864 questions). Overall F1 and its paired difference weight
documents equally; coverage columns weight questions equally. Covered answers
were elicited by the strict training probes (466 questions); uncovered answers
were not (398 questions). Intervals are paired document-cluster 95\% intervals.}
\label{tab:refinement}
\scriptsize\setlength{\tabcolsep}{2pt}
\begin{tabular}{@{}lrrrl@{}}
\toprule
Condition & Overall & Covered & Uncovered & $\Delta$ native [95\% CI] \\
\midrule
Native D2L & .742 & .846 & .626 & --- \\
ICL teacher & .807 & .910 & .683 & --- \\
\midrule
Rank 8 & .743 & .882 & .588 & $+.002$ [$-.021$, $+.024$] \\
Rank 16 & .746 & .888 & .587 & $+.004$ [$-.019$, $+.027$] \\
Rank 32 & .752 & .899 & .584 & $+.010$ [$-.013$, $+.033$] \\
Rank 64 & .738 & .903 & .553 & $-.003$ [$-.026$, $+.020$] \\
\bottomrule
\end{tabular}
\par\smallskip
\raggedright On the fitting probes, mean forward KL is .0105--.0111 and
generated-response agreement with the teacher is .977--.980 F1 across ranks.
These fitting diagnostics are distinct from the natural-question scores above.
\end{table}

\begin{table}[t]
\centering
\caption{Question-conditioned writes ($K=1$, Gemma test set).
The writer receives the evaluation question. Each paired contrast uses
its run's native baseline.}
\label{tab:conditioned}
\small\setlength{\tabcolsep}{3pt}
\begin{tabular}{@{}lrl@{}}
\toprule
Training & F1 & $\Delta$ native [95\% CI] \\
\midrule
Mixed-$K$ KL & .733 & $-.022$ [$-.034$, $-.010$] \\
$K=1$ KL & .742 & $-.014$ [$-.027$, $-.000$] \\
Contrastive KL & .727 & $-.026$ [$-.042$, $-.011$] \\
\bottomrule
\end{tabular}
\end{table}

\section{Details supporting RQ2}
\label{sec:numerical-results}

\subsection{Matched reader evaluations}
\label{sec:composition-tables}
Tables~\ref{tab:composition}--\ref{tab:confirmation} report the post-block
and write-site comparisons underlying Figure~\ref{fig:composition}.
The gated-reader results follow below. Paired contrasts weight documents
equally; absolute scores weight questions equally.

\begin{table*}[t]
\centering
\caption{Test set: question-mean F1 and answer containment (C), 1,127
questions per cell. Normalization uses $K^{-0.5}$. The reader is the post-block
mixed-$K$ reader, averaged over three seeds on Gemma and one on Qwen. ICL
containment is .831, .824, .820, .818, .786 on Gemma and .880, .877, .883, .892, .894
on Qwen at $K=1,2,4,8,16$.}
\label{tab:composition}
\small\setlength{\tabcolsep}{4pt}
\begin{tabular}{@{}lrrrrrrrrrrrr@{}}
\toprule
& \multicolumn{6}{c}{Gemma-2-2B} & \multicolumn{6}{c}{Qwen3-4B} \\
\cmidrule(lr){2-7}\cmidrule(l){8-13}
& \multicolumn{4}{c}{F1} & \multicolumn{2}{c}{C} & \multicolumn{4}{c}{F1} & \multicolumn{2}{c}{C} \\
$K$ & ICL & Native & Norm. & Reader & Native & Reader & ICL & Native & Norm. & Reader & Native & Reader \\
\midrule
1 & .873 & .755 & .755 & .761 & .677 & .689 & .764 & .336 & .336 & .723 & .683 & .634 \\
2 & .863 & .618 & .604 & .624 & .534 & .533 & .732 & .261 & .239 & .581 & .484 & .471 \\
4 & .858 & .445 & .421 & .465 & .328 & .339 & .702 & .194 & .178 & .462 & .321 & .343 \\
8 & .849 & .200 & .283 & .268 & .116 & .163 & .674 & .062 & .141 & .294 & .069 & .187 \\
16 & .825 & .026 & .229 & .117 & .002 & .049 & .661 & .016 & .115 & .155 & .000 & .073 \\
\bottomrule
\end{tabular}
\end{table*}

\begin{table*}[t]
\centering
\caption{Paired document-mean differences and 95\% source-cluster intervals on the
test set for the post-block mixed-$K$ reader. Qwen containment controls
for the response-format effect of gold-span training.}
\label{tab:contrasts}
\small\setlength{\tabcolsep}{5pt}
\begin{tabular}{@{}lrlll@{}}
\toprule
Model / metric & $K$ & Normalized $-$ native & Reader $-$ native & Reader $-$ normalized \\
\midrule
Gemma F1 & 1 & --- & $+.006$ [$-.008$, $+.020$] & --- \\
& 2 & $-.011$ [$-.024$, $+.002$] & $+.007$ [$-.009$, $+.024$] & $+.018$ [$+.001$, $+.035$] \\
& 4 & $-.026$ [$-.047$, $-.006$] & $+.016$ [$-.003$, $+.037$] & $+.042$ [$+.021$, $+.063$] \\
& 8 & $+.082$ [$+.060$, $+.104$] & $+.070$ [$+.053$, $+.088$] & $-.012$ [$-.032$, $+.008$] \\
& 16 & $+.206$ [$+.181$, $+.231$] & $+.093$ [$+.077$, $+.109$] & $-.113$ [$-.136$, $-.092$] \\
\midrule
Qwen containment & 1 & --- & $-.054$ [$-.080$, $-.029$] & --- \\
& 2 & $-.052$ [$-.072$, $-.033$] & $-.009$ [$-.039$, $+.020$] & $+.043$ [$+.013$, $+.073$] \\
& 4 & $-.054$ [$-.076$, $-.032$] & $+.023$ [$-.004$, $+.051$] & $+.078$ [$+.047$, $+.109$] \\
& 8 & $+.133$ [$+.107$, $+.159$] & $+.120$ [$+.096$, $+.144$] & $-.013$ [$-.040$, $+.013$] \\
& 16 & $+.147$ [$+.123$, $+.172$] & $+.073$ [$+.056$, $+.092$] & $-.074$ [$-.096$, $-.052$] \\
\bottomrule
\end{tabular}
\end{table*}

\begin{table*}[t]
\centering
\caption{Write-site deep rank reader on the test set (Gemma-2-2B; 1,127
questions; question-mean F1; three seeds) with seed-mean paired contrasts.
The reader was selected on development data and evaluated on the same frozen
test banks as the post-block reader.}
\label{tab:confirmation}
\small\setlength{\tabcolsep}{4pt}
\begin{tabular}{@{}lrll@{}}
\toprule
$K$ & Reader & Reader $-$ native & Reader $-$ normalized \\
\midrule
1 & .758 & $+.005$ [$-.010$, $+.019$] & --- \\
2 & .633 & --- & $+.028$ [$+.010$, $+.045$] \\
4 & .492 & $+.045$ & $+.071$ [$+.053$, $+.089$] \\
8 & .348 & $+.151$ & $+.069$ [$+.049$, $+.089$] \\
16 & .251 & $+.227$ & $+.021$ [$+.004$, $+.039$] \\
\bottomrule
\end{tabular}
\end{table*}

\subsection{Composition diagnostics}
\begin{table}[t]
\centering
\caption{Qwen gated reader on the matched test banks: 1,127 questions per $K$.
The $B$-row gate uses one seed and a validation-selected checkpoint.
Differences in containment
use paired document means and 95\% bootstrap intervals.}
\label{tab:qwen-gated-test}
\small\setlength{\tabcolsep}{3pt}
\begin{tabular}{@{}rrrl@{}}
\toprule
$K$ & F1 & Contains & Difference [95\% CI] \\
\midrule
1 & .431 & .619 & $-.066$ [$-.087$, $-.046$] \\
2 & .405 & .436 & $+.005$ [$-.022$, $+.035$] \\
4 & .337 & .272 & $+.007$ [$-.020$, $+.036$] \\
8 & .267 & .180 & $-.020$ [$-.042$, $+.002$] \\
16 & .183 & .105 & $-.042$ [$-.062$, $-.022$] \\
\bottomrule
\end{tabular}
\par\smallskip
\raggedright Reference: native D2L at $K=1$, normalization at $K\geq2$.
Cell scores are question means; 10,000 resamples cluster by source document.
\end{table}

Table~\ref{tab:qwen-gated-test} extends the gated-reader comparison to Qwen.
The $B$-row gated-reader configuration was fixed from Gemma; Qwen's checkpoint was selected by inline
validation macro F1, without test-based selection. This is an additional
evaluation on the previously used test set, with one seed.
F1 improves over native D2L at every $K$, but containment is lower at
$K=1,2,4$. Against normalization, containment intervals include zero at
$K=2,4,8$ and show a loss at $K=16$.

\begin{table}[t]
\centering
\caption{Gated reader on the Gemma test set: 1,127 questions per $K$.
The $B$-row gate uses one seed and a development-selected checkpoint.
F1 is question-mean; differences and 95\% intervals use paired document means
(10,000 source-cluster resamples).}
\label{tab:gated-test}
\small\setlength{\tabcolsep}{3pt}
\begin{tabular}{@{}rrl@{}}
\toprule
$K$ & F1 & Difference [95\% CI] \\
\midrule
1 & .750 & $-.005$ [$-.015$, $+.004$] \\
2 & .605 & $+.001$ [$-.015$, $+.017$] \\
4 & .418 & $-.001$ [$-.019$, $+.017$] \\
8 & .289 & $+.008$ [$-.011$, $+.027$] \\
16 & .184 & $-.045$ [$-.064$, $-.025$] \\
\bottomrule
\end{tabular}
\par\smallskip
\raggedright Differences are against native D2L at $K=1$ and
uniform normalization at $K\geq2$.
\end{table}

Table~\ref{tab:gated-test} adds the gated reader to the matched test comparison
in Figure~\ref{fig:composition}. The $B$-row gate was selected over the
latent-assisted gate using their mean $K=1,8$ development scores at the
validation-selected checkpoints. The test set had already
been used for the post-block and write-site readers. This additional evaluation
uses the same frozen random banks, with one gated-reader seed.

Table~\ref{tab:probes} gives the membership-probe measurements supporting
Section~\ref{sec:res-probes}.

\begin{table}[t]
\centering
\caption{Membership probes on the development set (Gemma-2-2B; held-out per-question
AUC at the document level, best layer in parentheses; chance 0.5). Query
state: the layer-normed MLP input at the stated positions. Clean: bank silent during
collection. All probes use $B$-row keys, matching Figure~\ref{fig:layer-probes}.}
\label{tab:probes}
\small\setlength{\tabcolsep}{4pt}
\begin{tabular}{@{}llrr@{}}
\toprule
Query state & $K$ & Best AUC & Layers 0--5 \\
\midrule
Response positions & 8 & .852 (8) & .55--.71 \\
Response positions & 16 & .743 (6) & .55--.67 \\
Prompt mean, clean & 8 & .821 (9) & .50--.71 \\
\bottomrule
\end{tabular}
\end{table}

\subsection{Layer-localization measurements}
Table~\ref{tab:layers} provides the values and paired differences for
Figure~\ref{fig:layers} and Table~\ref{tab:prompt-interventions}.
Absolute F1 averages questions; differences from the source-only baseline
average documents and use 10,000 source-cluster bootstrap resamples.

\begin{table*}[t]
\centering
\caption{Layer and prompt interventions. ``Native below $L$'' retains the
bank below $L$ and the source alone thereafter; ``silent below $L$'' removes
all document contributions below $L$. Differences are relative to native
$K=1$. Intervals are 95\%; a point estimate alone indicates an unreported
interval. Gemma/SQuAD and Qwen/SQuAD use 1,260 development questions;
DROP/ROPES uses 1,934 questions from 612 sources.}
\label{tab:layers}
\label{tab:layer-contrasts}
\label{tab:generality}
\small\setlength{\tabcolsep}{5pt}
\begin{tabular}{@{}llrrl@{}}
\toprule
Model / dataset & Intervention & $K$ & F1 & Difference from native $K=1$ \\
\midrule
Gemma / SQuAD & Native below 0 & 8 & .740 & $+.002$ [$-.001$, $+.007$] \\
& Native below 4 & 8 & .708 & $-.025$ [$-.043$, $-.008$] \\
& Native below 8 & 8 & .611 & $-.124$ [$-.149$, $-.098$] \\
& Native below 13 & 8 & .466 & $-.254$ [$-.288$, $-.218$] \\
& Native below 18 & 8 & .388 & $-.332$ \\
& Native below 22 & 8 & .234 & $-.492$ [$-.523$, $-.459$] \\
& Native below 0 & 16 & .742 & --- \\
& Native below 4 & 16 & .591 & $-.131$ [$-.157$, $-.104$] \\
& Native below 8 & 16 & .111 & $-.612$ \\
& Native below 13 & 16 & .035 & $-.698$ [$-.725$, $-.671$] \\
& Native below 18 & 16 & .024 & $-.710$ \\
& Native below 22 & 16 & .015 & $-.717$ [$-.744$, $-.690$] \\
& Silent below 4 & 8 & .724 & $-.011$ [$-.025$, $+.003$] \\
& Silent below 6 & 8 & .678 & $-.049$ [$-.066$, $-.032$] \\
& Silent below 8 & 8 & .633 & $-.103$ [$-.124$, $-.082$] \\
& Silent below 13 & 8 & .480 & $-.259$ [$-.290$, $-.227$] \\
& Source at last prompt and response positions & 8 & .437 & $-.294$ [$-.327$, $-.261$] \\
& Source applied with cache replay & 8 & .569 & $-.167$ [$-.194$, $-.139$] \\
& Distractors removed at response positions & 8 & .313 & $-.389$ [$-.424$, $-.351$] \\
& Also removed at last prompt position & 8 & .430 & $-.284$ [$-.317$, $-.249$] \\
\midrule
Gemma / DROP+ROPES & Native below 0 & 8 & .381 & $-.004$ [$-.008$, $+.001$] \\
& Native below 8 & 8 & .307 & $-.082$ [$-.102$, $-.063$] \\
& Native below 18 & 8 & .178 & $-.188$ [$-.210$, $-.165$] \\
& Silent below 6 & 8 & .359 & $-.027$ [$-.042$, $-.012$] \\
& Silent below 8 & 8 & .331 & $-.056$ [$-.074$, $-.039$] \\
\midrule
Qwen / SQuAD & Native below 0 & 8 & .265 & $+.001$ [$-.003$, $+.005$] \\
& Native below 11 & 8 & .181 & $-.082$ [$-.095$, $-.069$] \\
& Native below 25 & 8 & .041 & $-.229$ [$-.244$, $-.216$] \\
& Silent below 8 & 8 & .253 & $-.014$ [$-.023$, $-.006$] \\
& Silent below 11 & 8 & .239 & $-.030$ [$-.039$, $-.020$] \\
\bottomrule
\end{tabular}
\end{table*}

Native $K=1$ F1 is .740 on Gemma/SQuAD, .385 on Gemma/DROP+ROPES,
and .264 on Qwen/SQuAD. Relative-depth comparisons use layers 8 and 18
of Gemma's 26 layers and layers 11 and 25 of Qwen's 36 layers.
The loss at one-third depth is 0.36--0.37 of the loss at two-thirds depth
in all three evaluations. This replicates the qualitative pattern but does
not meet the preregistered threshold of 0.5; the Gemma/SQuAD evaluation
used to formulate that threshold also falls below it.

\section{Details supporting RQ3}
\subsection{PersonaMem-v2}
Table~\ref{tab:personamem} gives paired accuracy differences for the
conditions in Table~\ref{tab:user-memory-summary}. Each question has the
same retrieved episodes across the episodic conditions.
\begin{table}[t]
\centering
\caption{PersonaMem-v2 paired accuracy differences (195 personas,
4,878 questions). Intervals use 4,000 persona-cluster bootstrap resamples
with question-weighted estimates. Adapter comparisons retain episodic text.}
\label{tab:personamem}
\small\setlength{\tabcolsep}{3pt}
\begin{tabular}{@{}lrl@{}}
\toprule
Comparison & Difference & 95\% CI \\
\midrule
Both text $-$ episodic only & $+.0133$ & [$+.0047$, $+.0219$] \\
Summed $-$ joint LoRA & $-.1169$ & [$-.1355$, $-.0968$] \\
Normalized $-$ joint LoRA & $-.0041$ & [$-.0129$, $+.0049$] \\
\bottomrule
\end{tabular}
\end{table}

\subsection{Serving}
Tables~\ref{tab:cost-length} and~\ref{tab:throughput} give the accounting
and timing measurements underlying Figure~\ref{fig:serving-tradeoff}.
\begin{table}[t]
\centering
\caption{Gemma-2-2B analytical serving cost: one document, 40 question tokens,
32 generated tokens. D2L state counts the per-user document adapter in bfloat16;
the global shared bias adds 9.6\,MB once. Per-request MACs include both adapters
but exclude compilation.}
\label{tab:cost-length}
\small\setlength{\tabcolsep}{3pt}
\begin{tabular}{@{}rrrrr@{}}
\toprule
& \multicolumn{2}{c}{GMAC/request} & \multicolumn{2}{c}{State (MB)} \\
\cmidrule(lr){2-3}\cmidrule(l){4-5}
$L$ & ICL & D2L & KV & D2L \\
\midrule
150 & 583.0 & 188.8 & 16.0 & 9.6 \\
512 & 1{,}544.8 & 188.8 & 54.5 & 9.6 \\
1{,}024 & 2{,}929.1 & 188.8 & 109.1 & 9.6 \\
2{,}048 & 5{,}781.3 & 188.8 & 218.1 & 9.6 \\
4{,}096 & 11{,}820.7 & 188.8 & 436.2 & 9.6 \\
\bottomrule
\end{tabular}
\end{table}

\begin{table}[t]
\centering
\caption{Batched serving on one A100, four documents per user.
Throughput includes prefill and decoding of 64 tokens per user.
Peak memory is measured at batch size 16. Compilation is excluded;
ICL performs full prefill for each request.}
\label{tab:throughput}
\small\setlength{\tabcolsep}{4pt}
\begin{tabular}{@{}llrrrr@{}}
\toprule
Model & Method & Batch 1 & Batch 8 & Batch 16 & Peak GB \\
\midrule
Gemma & ICL & 24.3 & 165.3 & 290.9 & 21.9 \\
 & D2L & 20.1 & 157.8 & 315.8 & 18.9 \\
Qwen & ICL & 21.5 & 122.0 & 162.5 & 35.9 \\
 & D2L & 17.1 & 134.9 & 269.4 & 22.0 \\
\bottomrule
\end{tabular}
\end{table}

\clearpage

\end{document}